\documentclass[preprint,12pt]{elsarticle}

\usepackage{amssymb}
\usepackage{textcomp, gensymb}
\usepackage{amsmath}
\usepackage{booktabs}
\usepackage{accents}
\usepackage{bm}

\makeatletter
\newif\ifnopreprintline
\nopreprintlinetrue
\makeatother
\begin{document}




\title{R-FORCE: Robust Learning for Random Recurrent Neural Networks}

\author{Yang Zheng\fnref{aff1}}
\author{Eli Shlizerman\corref{cor1}\fnref{aff1,aff2}}

\cortext[cor1]{Corresponding author}

\ead{shlizee@uw.edu}

\affiliation[aff1]{organization={Department of Electrical and Computer Engineering, University of Washington},
addressline={185 Stevens Way, Paul Allen Center}, 
city={Seattle},
postcode={98195-2500}, 
state={WA},
country={US}}

\affiliation[aff2]{organization={Department of Applied Mathematics, University of Washington},
addressline={185 Stevens Way, Paul Allen Center}, 
city={Seattle},
postcode={98195-2500}, 
state={WA},
country={US}}

\begin{abstract}
Random Recurrent Neural Networks (RRNNs) offer a straightforward approach to model and extract features from sequential data, yet they are prone to the diminishing/exploding gradient problem during gradient-descent-based optimization. To address this limitation, an alternative training approach, called FORCE learning, proposes a recursive-least-square-based training for RRNNs. It enables RRNNs to effectively handle demanding tasks such as target learning, wherein networks autonomously generate complex sequential patterns over extended periods without external guiding input. While FORCE learning shows promise, its effectiveness has been predominantly constrained to specific dynamic regimes, typically near the edge-of-chaos.

To extend the application of FORCE learning, we conduct a thorough analysis of the network Jacobian matrix, identifying key features exclusive to successful dynamics. Recognizing the pivotal role of the recurrent connectivity matrix on the Jacobian spectrum, we propose a novel initialization strategy for this connectivity matrix designed to impose desired features on the Jacobian matrix over a broad regime of dynamics. Specifically, this initialization method follows four generating principles, ensuring the eigenvalues of the resulting Jacobian matrix remain within the the stability region. Integrating this initialization approach with FORCE learning, we introduce Robust-FORCE (R-FORCE), an approach robust to diverse network dynamics. We validate R-FORCE through applications involving the generation of multidimensional signals such as human body movements, demonstrating superior accuracy and stability compared to standard FORCE learning and other SOTA methods. Furthermore, we illustrate the versatility of R-FORCE by successfully applying it to biologically plausible systems, i.e., spiking neural networks, including Theta and Izhikevich neurons. Our experiments, involving tasks such as sine-wave generation and birdsong spectrogram reproduction, confirm that R-FORCE consistently outperforms its FORCE counterpart in terms of accuracy and reliability.
\end{abstract}


\maketitle

\section{Introduction}
Recurrent neural networks (RNNs) have proven effective in learning features from multi-dimensional sequential data, enabling their successful application in various machine learning tasks such as speech recognition, music composition, and human action modeling or recognition~\cite{yao2014spoken, graves2013speech, eck2002learning, shlizerman2018audio, baccouche2011sequential, su2020clustering}. While sequence types vary across applications, the underlying nature of training RNN is gradient-descent-based optimization, which introduces the infamous diminishing/exploding gradient problem. While gated RNNs (e.g., LSTM~\cite{hochreiter1997long}, GRU~\cite{cho2014learning}) alleviate this issue, they encounter difficulties in fundamental tasks such as \textit{target-learning}, where the network \textit{generates} a prescribed sequence without input for thousands of time steps. Alternative architectures and training approaches have been proposed to tackle this narrow choice of network architectures and the incompleteness of the target-learning task. Furthermore, it is intriguing to find out if these alternative architectures and training approaches are biologically plausible and could inherit the effectiveness of brain networks in processing multi-dimensional sequences.

Random RNNs (RRNNs) have gained attention due to their simplicity, and various training approaches have been applied to them to tackle the target-learning task. Echo State~\cite{jaeger2001echo}, during training, clamps the network to achieve a feed-forward propagation to avoid the delay effects and chaotic dynamics in RRNNS. However, it is sensitive to perturbations and requires extended training~\cite{buonomano1995temporal, maass2002real, jaeger2003adaptive}. FORCE learning offers an alternative by employing a recursive-least-square-based optimization, rapidly dampening errors through a feedback loop~\cite{sussillo2009generating}. FORCE learning successfully addresses the target-learning task in configurations, such as Rank1-RRNN and RankN-RRNN, where a single read-out unit aggregates information from neural populations. These configurations are biologically plausible, reflecting similar neural network structures observed in the brain. However, training Rank1-RRNN is challenging because only the weights connecting the read-out unit and the hidden states are trainable, constraining the optimization to a single dimension.

Despite the success achieved with various training approaches, RRNNs exhibit inherent instability in the target-learning task. This instability arises because RRNNs are either in quiescent or chaotic states for most parameter settings drawn from normal distributions. These states are considered ”non-trainable” because training approaches must either prevent energy dissipation or control chaotic behavior~\cite{sompolinsky1988chaos}. Only for a few parameter settings might the network be in a transitional state between these two states (i.e., edge-of-chaos). Specifically, for Rank1- and RankN-RRNN, FORCE learning demonstrates optimal performance when the network, prior to training, is situated in the edge-of-chaos regime. This is characterized by setting the global scaling parameter of the connectivity matrix, denoted as $g$, within the range of $[1.3,1.5]$. In this regime, the achieved low error during training remains controlled for thousands of time steps during testing. However, even within this regime, performance is not flawless. Approximately $15\%$ of network initialization may result in sequences diverging from the target under the most optimal setup. A recent adaptation of FORCE, called Full-FORCE, managed to reduce this failure rate to approximately $7\%$ by incorporating an auxiliary RRNN~\cite{depasquale2018full}. While Full-FORCE enhances network robustness to dynamic chaos, it demands extended training and cannot entirely eliminate failures. This observation motivates our exploration of methods that can produce more reliable generation, which can be achieved by a stabilized training process~\cite{hardt2016train}.

Various strategies have been explored to stabilize gradient-descent-based optimization in RRNNs. Direct methods, such as gradient clipping and normalization, explicitly modify the gradients during optimization~\cite{pascanu2013difficulty, chen2017gradnorm, graves2013generating}. Indirect methods impose additional constraints on the connectivity matrix to facilitate optimal training and inference. A significant subset of these approaches involves constraints on the initialization of the connectivity matrix, where specific features, such as identity, orthogonal, or antisymmetric matrices, are employed~\cite{mhammedi2017efficient, sussillo2009generating, mikolov2014learning, le2015simple, maduranga2019complex, wisdom2016full, arjovsky2016unitary, helfrich2017orthogonal, vorontsov2017orthogonality, chang2019antisymmetricrnn}. The influence of the connectivity matrix on training stability is notably reflected in the network Jacobian matrix. This matrix quantifies the correlation between the current states and initial states of the network, providing crucial insights into the network's characteristics. Specifically, the positions of Jacobian matrix eigenvalues (spectrum) in the complex plane serve as indicators of potential network stability or instability~\cite{haber2017stable, laurent2016recurrent}. Eigenvalues outside the stability region can introduce instability, while eigenvalues within the stability region may cause rapid information decay. Consequently, it is recommended to position the eigenvalues close to the boundary of the stability region to prevent instability and rapid decay. For example, Antisymmetric RNN employs an initialization strategy that keeps the Jacobian spectrum in the stable region. This involves setting the initial connectivity as an antisymmetric matrix with a diffusion term, ensuring that all eigenvalues are distributed in a narrow band near the imaginary axis~\cite{chang2019antisymmetricrnn}. Although effective for traditional gradient-based training, such initialization strategies are not directly applicable to FORCE learning due to differences in the training approach. This motivates our investigation into optimal connectivity initialization tailored specifically for FORCE learning.

To delve into the factors contributing to the success of FORCE learning within a specific regime, we conduct an in-depth analysis of the spectral properties of the Jacobian matrix. Building on this analysis, we derive four generating principles for generating the spectrum of the connectivity matrix, ensuring the corresponding Jacobian matrix possesses these essential spectral properties. We propose Robust-FORCE (R-FORCE), which initializes the connectivity matrix according to these principles and then applies it to FORCE learning. This initialization facilitates stable evolution and efficient training. We initially evaluate R-FORCE on various synthetic one-dimensional targets, including discontinuous and noisy signals, to demonstrate its robustness across network dynamics. Additionally, we apply R-FORCE to configurations such as Rank1- and RankN-RRNN. In terms of real-world data, R-FORCE is validated using Physical Rehabilitation Movements dataset (UI-PRMD)~\cite{vakanski2018data}, characterized by multi-dimensional targets. R-FORCE achieves lower error and tighter confidence intervals than FORCE learning and other SOTA methods across diverse network dynamics. This reliability is crucial for multi-dimensional tasks, where outliers increasingly degrade overall quality with target dimensionality. Highlighting biological plausibility, we extend R-FORCE to spiking neural networks, specifically Theta and Izhikevich neurons. Experimental results, including simple sine wave generation and complex 33-band-spectrogram birdsong reproduction, demonstrate superior accuracy and reliability compared to standard FORCE methods.

\begin{figure}[t]
    \centering
    \includegraphics[width=\linewidth]{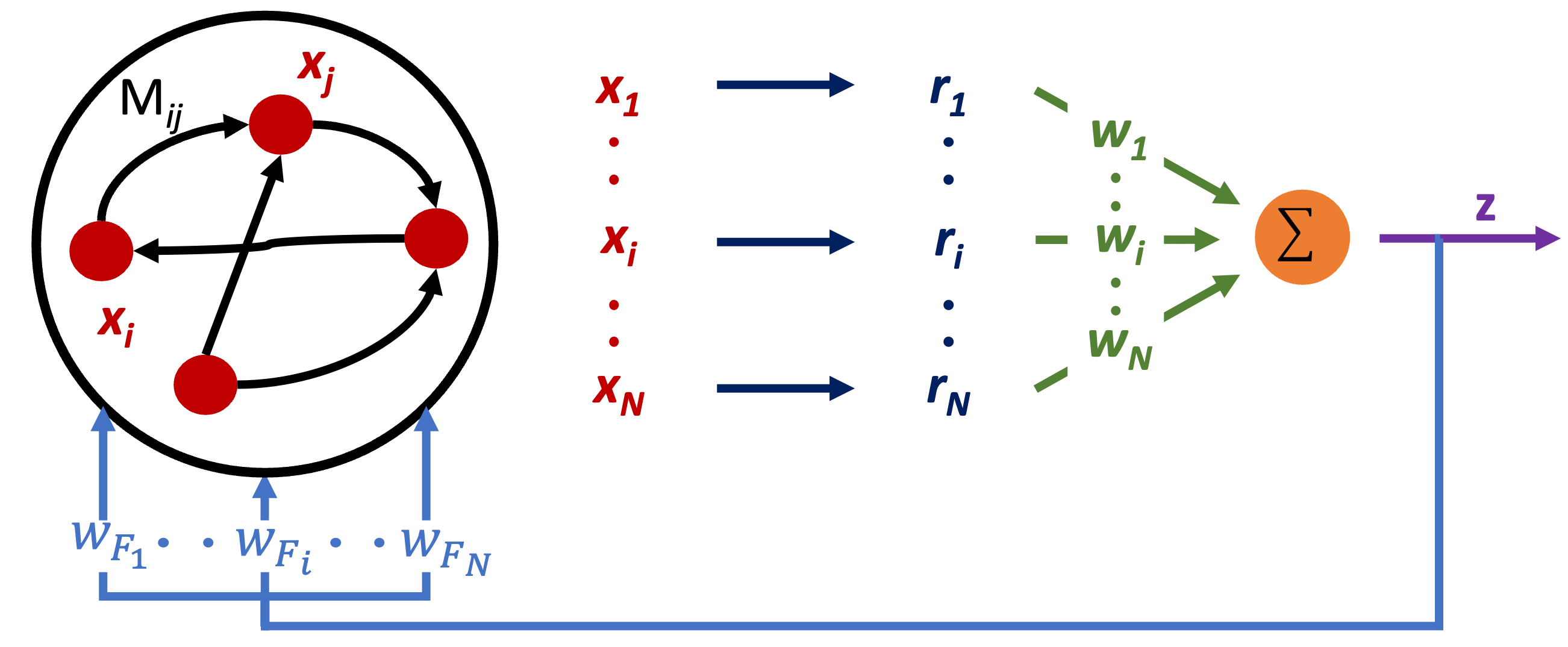}
    \caption{\textbf{Rank1-RRNN.} A reservoir of neurons $x_i$ (red) interconnected through a randomly initialized connectivity matrix $M$ (black). Neurons are activated by $\tanh$ ($r_i$ units (navy)) and connected to a read-out unit $z$ (purple) by $\bm{w}$ (green). The network also includes a fixed feedback loop from the read-out unit to the reservoir ($\bm{w_F}$ (blue)). During training, only the vector $\bm{w}$ is modified.}
    \label{fig:networksetup}
\end{figure}

\section{Methods}
\subsection{Spectral Analysis of FORCE Learning}\label{sec:SpecR1-RNN_FORCE}

Rank1-RRNN, as illustrated in Fig.~\ref{fig:networksetup}, is the simplest RRNN configuration proposed for FORCE learning~\cite{sussillo2009generating}. It is a firing-rate model, and its evolution is described by the following Eq.~\ref{eq:firingRate}-\ref{eq:linearOutput}:

\begin{align}
    \tau \frac{dx_i}{dt} & = -x_i + g \sum^N_{j=1} M_{ij} r_j + {w_F}_i z,
    \label{eq:firingRate} \\
    \bm{r(t)} & = \tanh(\bm{x(t)}), 
    \label{eq:activation}\\
    z(t) & = \bm{w(t)} ^T \bm{r(t)}.
    \label{eq:linearOutput}
\end{align}
$\textbf{x}$ $\in \mathcal{R}^{N}$ represents the state of neurons, where $N$ is the number of neurons. $\tau$ is the basic time constant in the model. Neurons are interconnected through the connectivity matrix $M \in \mathcal{R}^{N \times N}$, whose weights are scaled by a single parameter $g$. In Rank1-RRNN, these weights are fixed and are randomly sampled from a normal distribution $\mathcal{N}(0,\frac{1}{\sqrt{p N}})$, with $p \in [0,1]$ indicating the sparsity of $M$. Neurons are activated through a nonlinear function $tanh$ and linearly integrated into a read-out unit $z$ via the 
vector $\bm{w}$. Additionally, the read-out unit $z$ has a feedback connection to the neurons through a column vector \bm{$w_F$}. The elements of \bm{$w_F$} are sampled from a uniform distribution $U(-1,1)$ and are fixed over training. In summary, after the Rank1-RRNN is initialized, only vector \bm{$w$}, connecting the activated units $\textbf{r}$ and read-out unit $z$, is optimized during training. For more details, see~\cite{sussillo2009generating} and references therein.

\begin{figure*}[t]
\includegraphics[width=1\linewidth]{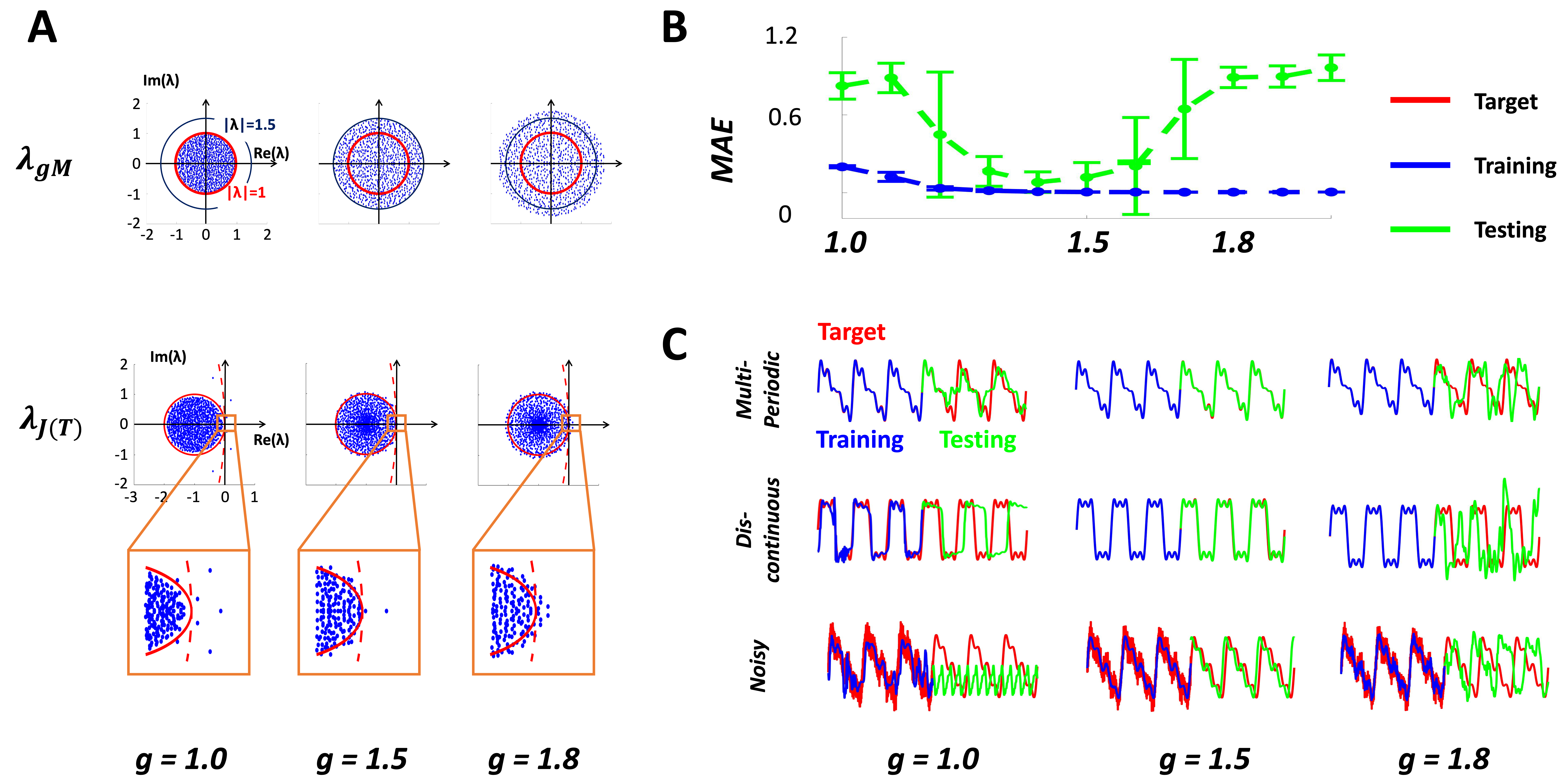}
\caption{
 \textbf{FORCE Learning with Rank1-RRNN.} (A) Spectrum of the scaled connectivity matrix $gM$ (top) and the end-of-training Jacobian matrix $J(T)$ (bottom) for $g=1.0, 1.5, 1.8$. (B) Training (blue) and testing (green) performance over the range of $g \in [1,2]$. (C) Comparison between target sequences (red) and the sequences generated during training (blue) and testing (green).
} 
\label{fig:FORCE_Results}
\end{figure*}

Upon initializing the connectivity matrix $M$, its spectrum appears to be distributed homogeneously within the unit circle in the complex plane, as discussed in~\cite{vakanski2018data}. Accordingly, the eigenvalues of the scaled matrix $gM$ are distributed within the circle of radius $g$, as shown in the first row of Fig.~\ref{fig:FORCE_Results}~A. By adding additional connections to and from the read-out unit $z$, we obtain the extended matrix $\tilde{M} \in \mathcal{R}^{(N+1) \times (N+1)}$, i.e., additional column vector \bm{$w_F$} and row vector \bm{$w^T$}.
\begin{equation*}
\tilde{M} = 
\left( \begin{matrix}
M & \bm{w_F} \\
\bm{w^T} & 0
\end{matrix}
\right).
\end{equation*}

Notably, $\tilde M$ appears to remain similar over training. This is expected given that only one single dimension of the matrix, i.e., $\bm{w}$, is being updated~\cite{bunch1978rank}.

Inspired by previous work~\cite{chang2019antisymmetricrnn}, we compute $J(t)$, i.e., the Jacobian matrix of the current state \bm{$x(t)$} with respect to the initial state \bm{$x(0)$}, to better estimate the training stability of the network. The Jacobian $J(t)\in \mathcal{R} ^{N\times N}$ is defined as follows: 
\begin{equation}
    J(t) = - \mathcal{I} + (gM + \bm{w_F} \bm{w(t)^T}) \times Diag(\bm{1}  - \tanh(\bm{x(t)}^2),
    \label{eq:Jacobian}
\end{equation}
where $\mathcal{I}$ represents the identity matrix, and $Diag(\cdot)$ indicates a diagonal matrix constructed from the vector argument. \bm{$1$} is a column vector of ones. After calculating $J(t)$, we investigate the location of its eigenvalues (spectrum) relative to the stability region defined by the Forward Euler method, which is applied to evolve the dynamics in Eq.~\ref{eq:firingRate}. Specifically, the stability region is a circle in the complex plane with its center at $-\tau$ and radius $\tau$. Eigenvalues within the stability region mean the evolution is contracting/vanishing, while eigenvalues outside the stability region indicate instability. Therefore, it is desirable to place most of the eigenvalues within the stability region and keep the remaining ones close to the boundary to ensure non-vanishing evolution.

The second row of Fig.~\ref{fig:FORCE_Results}~A illustrates the end-of-training Jacobian spectra $\lambda_{J(T)}$ for different scaling parameters $g$. Each plot shows eigenvalues (blue dots), the unit circle (solid red line), and the boundary of the stability region 
with $\tau=0.1$ (dashed red line).Detailed zoom-ins highlight that for extreme values such as $g = 1.0$ and $g=1.8$, eigenvalues fall predominantly either insider or outside the boundary, indicating a lossy or unstable system, respectively. Conversely, for intermediate values (e.g., $g = 1.5$), eigenvalues align near the boundary, suggesting balanced dynamics and optimal conditions for long-term dependency learning~\cite{chang2019antisymmetricrnn}. These observations align with the notion that FORCE learning is successful exclusively in the edge-of-chaos regime ($g=1.5$)~\cite{sussillo2009generating}.

To substantiate this finding and investigate the sensitivity to the scaling parameter $g$, we evaluate network performance on a four-sine-wave target function, as performed in \cite{sussillo2009generating}. We train the network with uniformly spaced $g$ values in the range $g \in [1,2]$ with a step of $0.1$ and $8$ repeated trials for each $g$. Fig.~\ref{fig:FORCE_Results}~B shows the corresponding Mean Absolute Error (MAE) for each $g$, where $MAE=\sum_{t=1}^T |f(t) - z(t)|/T$. The result indicates that the FORCE learning is sensitive and performs optimally in a narrow range of $g$ values, namely $g_{opt} \in [1.3, 1.5]$. The narrow performance regime appears consistent across target functions. Fig.~\ref{fig:FORCE_Results}~C demonstrates the performance of Rank1-RRNN with three different target functions: (i) a superposition of five sine waves (multi-periodic), (ii) a rapidly changing superposition (discontinuous), and (iii) a superposition of four sine waves with Gaussian noise (noisy). For $g < g_{opt}$ (e.g., $g=1.0$), while the network's generations typically follow the target during training, it eventually dissipates information and loses the learned pattern during testing. In contrast, for $g > g_{opt}$ (e.g., $g=1.8$), the network becomes sensitive to small perturbations during testing, causing the error to snowball and the generated sequence to diverge from the ground truth target. These experimental results align with the $J(T)$ spectrum distribution interpretation in Fig.~\ref{fig:FORCE_Results}~A. Moreover, even within the narrow range of $g_{opt}$ (e.g., $g = 1.5$), the network still fails to effectively learn patterns and make consistent predictions in some cases (e.g., noisy target).

Given these results, we explore alternative initialization methods to enhance stability and robustness over a broader range of scaling parameter $g$. Consequently, we introduce \textbf{Robust FORCE (R-FORCE)}, which initializes the connectivity matrix $M$ to ensure desirable Jacobian spectral characteristics, thus extending the effective application range of FORCE learning.

\subsection{R-FORCE Initialization} 

As shown in Eq.~\ref{eq:Jacobian}, the Jacobian $J(t)$ is determined by four different components: \bm{$w(t)$}, \bm{$x(t)$}, \bm{$w_F$}, and $M$. Among these, the first three are vectors and exert a minor (rank-1) influence on $J(t)$, while the matrix $M$ has a more significant (rank-N) impact~\cite{bunch1978rank}. Empirical evidence confirms that during training and testing, $J(t)$ shares a similar spectrum as $M$. Therefore, our focus lies on the initialization of $M$, aiming to position $\lambda_{J(t)}$ in the stability region where they hover around at all times.

We introduce a new initialization for $M$, termed $M_R$. To generate $M_R$, we first generate its eigenvectors and eigenvalues, and then apply the eigendecomposition, shown in Eq.~\ref{eq:generatingM_R}.

\begin{equation}
    M_R = V_R \times D_R \times V_R^{T},
    \label{eq:generatingM_R}
\end{equation}
where $V_R$ is an orthogonal matrix, i.e., $V_R \times V_R^{T} = I$, and its columns represent eigenvectors. $D_R$ is a diagonal matrix containing the corresponding eigenvalues.

\textbf{Generation of $M_R$ Eigenvectors}. To ensure $M_R$ real-valued, $V_R$ must be orthogonal and include complex conjugate column pairs~\cite{van1983matrix}. We first generate a real-valued random matrix $M_N \in \mathcal{R}^ {N \times N}$ from a normal distribution $\mathcal{N}(0,1)$ and construct an antisymmetric matrix $M_A = M_N - M_N^{T}$. The eigendecomposition of $M_A$ yields conjugate eigenvectors, formiing $V_R$.

\begin{figure*}[t]
    \centering
    \includegraphics[width=0.9\linewidth]{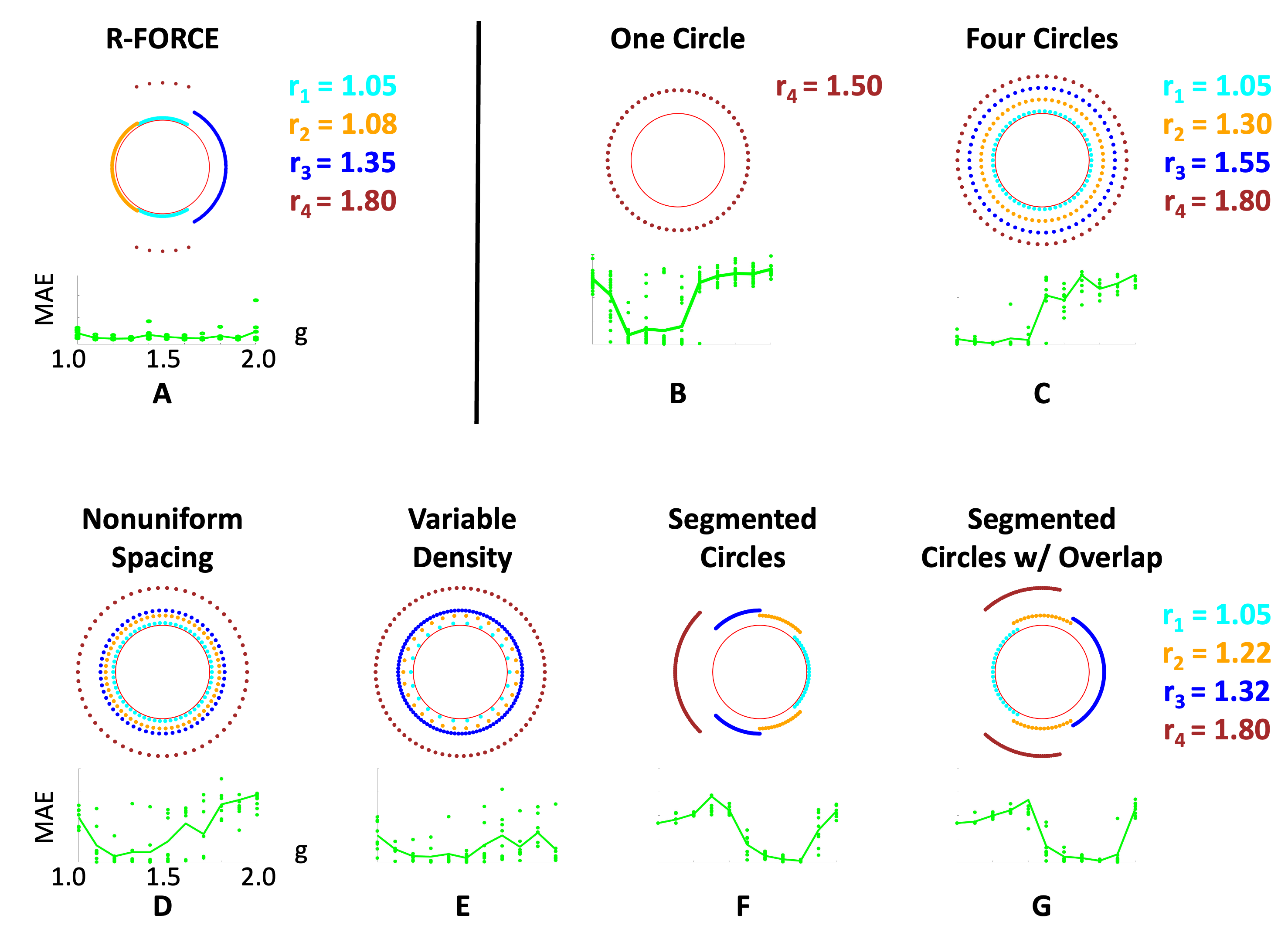}
    \caption{\bm{$\lambda_{M_R}$}\textbf{ and other candidate distributions}. (A-G): top row is the corresponding distribution of $g=1.5$ where $r_{1 \sim 4}$ denote the radii of circles. Configurations D-G have the same radii. The bottom row is the testing error curve of the given distribution over $g \in [1,2]$. For each configuration: (A) R-FORCE (optimal distribution) - Four arc segments with appropriate overlap, nonuniform radii, and variable density of eigenvalues (B) One circle with radius $g$, (C) Four circles with uniform spacing (D) Four circles with nonuniform spacing, (E) Four circles with nonuniform spacing and variable density, (F) Four arc segments with nonuniform spacing and variable density (G) Four arc segments with overlap (inappropriate), nonuniform spacing and variable density.}
    \label{fig:R-FORCE_otherDistributions}
\end{figure*}

\textbf{Generation of $M_R$ Eigenvalues}.
We seek an optimal distribution for $\lambda_{M_R}$ with that $\lambda_{J(t)}$ are kept within the stability region for all $t$. Our exploration of various configurations led us to identify four essential properties for the distribution. Our method, R-FORCE, initializes the connectivity matrix ${M_R}$ following those properties, and the spectral distribution of ${M_R}$ is shown in Fig.~\ref{fig:R-FORCE_otherDistributions}~A. Notably, since the generation of $M_R$ involves random terms, R-FORCE initialization can generate many stable variants of $M_R$.

To validate R-FORCE, we visualize and assess each of the four design principles (Fig.~\ref{fig:R-FORCE_otherDistributions}~B-G) on a common target-learning task composed of four sine waves. Each eigenvalue distribution is evaluated over scaling parameter $g \in [g_{min},g_{max}]=[1,2]$ using a Rank1-RRNN with $N = 1000$ neurons. Since $M_R$ is real-valued, only half the eigenvalue ($500$) are generated, and the other half are their conjugates.

\textit{Circular Distribution}.
The magnitude of eigenvalues plays a crucial role in determining system stability. A circular distribution centered at the origin is applied to ensure $\lambda_{M_R}$ possesses the same magnitude. Specifically, with $\lambda_{M_R}= r e^{i \theta}$, eigenvalues will have the same magnitude and various combinations of oscillatory (imaginary) and dissipative/expanding (real) components. This distribution is tuned by two parameters: $r$, and $\theta$. We set $r = g$ and uniformly sample $\theta$ between $[0, \pi]$. Although this distribution seems theoretically promising, its error is controlled only in a narrow range of $g$ values ($g \in [1.2, 1.4]$), similar to the vanilla FORCE learning, as shown in Fig.~\ref{fig:R-FORCE_otherDistributions}~B.
To extend the effective range of $g$ and keep the simplicity of circular distribution, we include more circles with different radii to the distribution.

\textit{Sets of Circles}. We generate $m$ concentric circles, where the radii of the largest and smallest circles are $1.2\times g/g_{min}$ and $1.4\times g/g_{max}$, respectively. The choice of those values is motivated by the observation that the optimal range of $g$ 
in the one-circle distribution is $[1.2,1.4]$. Therefore, this arrangement of the largest and smallest circles covers the whole $g$ range $(g\in[1, 2])$. Except for those two circles, the remaining $(m-2)$ middle circles are uniformly spaced between them. As $m$ increases, performance improves, but it plateaus for distributions with $m \geq 4$. Fig.~\ref{fig:R-FORCE_otherDistributions}~C shows the result obtained by $4$ circles. Therefore, we set $m=4$ for further distribution extensions. While this approach significantly improves performance for low $g$ values ($g \leq 1.4$), it does not yield the same performance for $g > 1.4$. We, therefore, implement non-uniformly distributed circles to explore the impact of middle circles. Fig.~\ref{fig:R-FORCE_otherDistributions}~D demonstrates a typical example of such distribution. It leads to a more robust testing error curve for $g > 1.4$ but is no longer optimal for $g \leq 1.4$.


\textit{Density}. Inspired by the observation that an eigenvalue with a large magnitude is more prone to instability, and only a few of these should be included, we apply different eigenvalue densities ($d_i$) to each circle. Specifically, the density is inversely proportional to the radius. As depicted in Fig.~\ref{fig:R-FORCE_otherDistributions}~E, it improves the performance in terms of average error and confidence interval for all examined $g$ values. However, such a distribution fails to yield superior performance when focusing on $g=1.5$ (the optimal scenario for FORCE learning).

\textit{Segmented Circles}.
Another plausible distribution variant applies arc segments instead of full circles, which can be achieved by generating $\theta$ in a specific range rather than $[0, \pi]$. Due to the conjugate pairs of the eigenvalues, the arcs will be replicated on their conjugated quadrant. We first equally divide the upper half of the complex plane into four fan-shaped regions without overlap, each containing one arc with a radius $r_i$. Eigenvalues are then uniformly sampled on each arc, following the same densities as Fig.~\ref{fig:R-FORCE_otherDistributions}~E. The segmented configuration, shown in Fig.~\ref{fig:R-FORCE_otherDistributions}~F, results in an error curve with optimal values at higher $g$ values $(g \geq 1.6)$. Notably, when $g \geq 1.6$, the rightmost arc ($r_1$) has a radius close to $[1.2, 1.4]$, consistent with the single circle case in Fig.~\ref{fig:R-FORCE_otherDistributions}~B. This observation suggests that positioning the right component of the $M_R$ spectrum into the optimal region $[1.2, 1.4]$ is crucial for network stability. These insights lead to the development of the R-FORCE distribution in Fig.~\ref{fig:R-FORCE_otherDistributions}~A, which combines the four generating principles mentioned above: multiple circular sets, non-uniform distance, unequal density, and segmented sets. As a result,
the R-FORCE distribution significantly improves performance and provides stable configurations for Rank1-RRNN. The radius and density of each circle follow an empirically obtained formula:
\begin{align}
\centering
    l_i & = \frac{g^2}{|r_i - 1.15|};
    \label{eq:eigPartition_1}\\
    d_i & = Q \times \frac{l_i}{\sum_{j=1}^k l_j} \hspace{0.5cm} i = 1, 2, \dots, k.
    \label{eq:eigPartition_2}
\end{align}

where
\begin{align*}
[r_1, r_2, r_3, r_4] & = [0.7, 0.72, 0.9, 1.2] \times g, \\
[Q,~~k] & = 
\begin{cases}
    [100\%, ~~m],& \text{if } r_{max} \leq 1.55 \\
    [99\%,~~m-1],           & \text{otherwise}
\end{cases}
\end{align*}

\begin{align*}
\theta_1, \theta_2, \theta_3, \theta_4 = 
\begin{cases}
    [72\degree, 144\degree], [144\degree, 180\degree], [0\degree, 72\degree], [0\degree, 72\degree], & \text{if } g < 1.4 \\
    [72\degree, 144\degree], [0\degree, 72\degree],[144\degree, 180\degree],  [72\degree, 144\degree], & \text{if } g > 1.8 \\
    [72\degree, 144\degree], [144\degree, 180\degree], [0\degree, 72\degree], [72\degree, 144\degree], & \text{otherwise}
\end{cases}
\end{align*}
$d_i$ is the percentage of eigenvalues placed on the $i$-th circle with the radius $r_i$. We also implement a threshold for the radius of the largest circle to determine $Q$ and $k$. When the largest radius exceeds this threshold, we only place $1\%$ of eigenvalues on this circle and distribute the other $99\%$ following $d_i$. Otherwise, all eigenvalues are assigned following $d_i$. When $g < 1.8$, $\theta_1 \in[72\degree, 144\degree]$, $\theta_2 \in[144\degree, 180\degree]$ and $\theta_3 \in[0\degree, 72\degree]$. Otherwise, the locations of $\theta_2$ and $\theta_3$ are swapped. 
The circle $r_4$ overlaps with $r_3$ for low $g$ values ($g < 1.4$) and otherwise overlaps with $r_1$.

To verify the necessity of moving circles, Fig.~\ref{fig:R-FORCE_otherDistributions}~G shows another setup where the angular positions of four arcs are fixed and $r_4$ overlaps with $r_2$ for all $g$. The testing error is controlled only when the right components of the spectrum are in the optimal range. This indicates that moving circles to keep the proper right component of the spectrum is crucial for the success of stable training.

\subsection{R-FORCE Properties}
\begin{figure*}[t]
    \centering
    \includegraphics[width=1\linewidth]{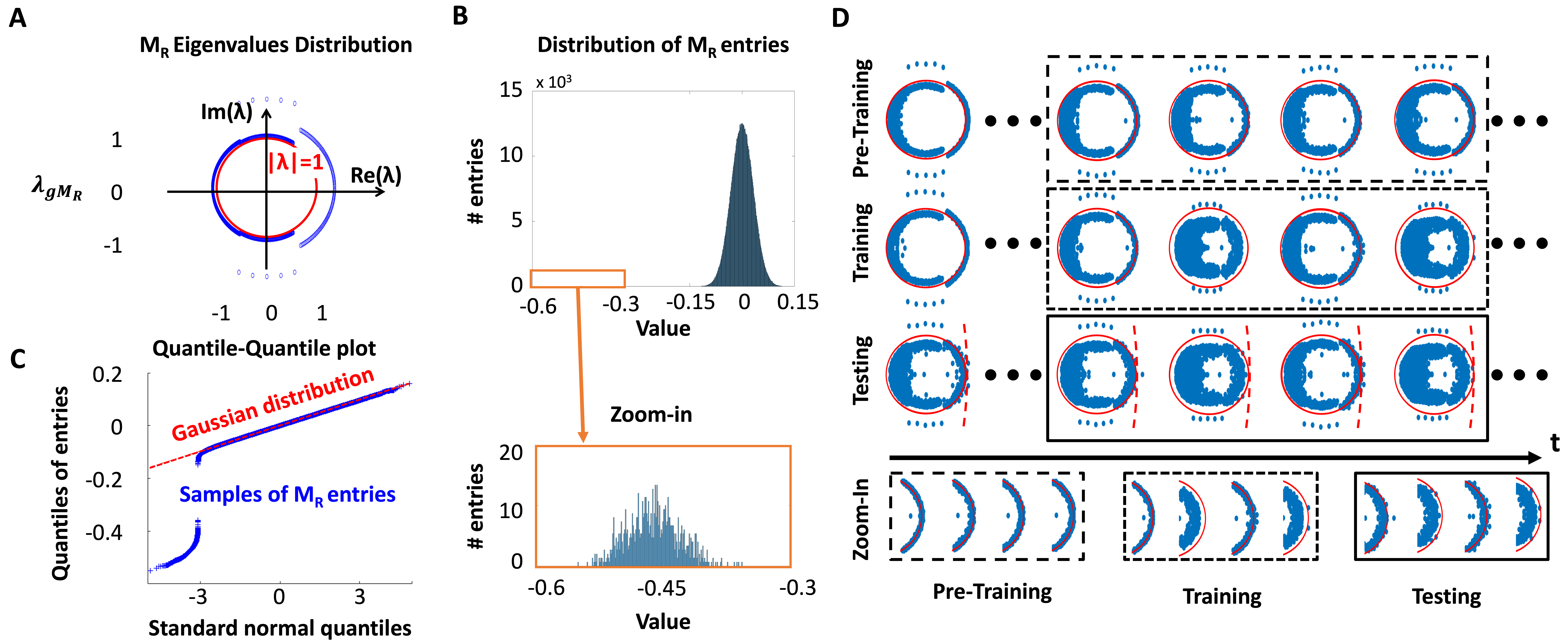}
    \caption{\textbf{R-FORCE Initialization Properties.} (A) Spectral distribution of $M_R$. (B) Histogram of $M_R$ elements. (C) Q-Q plot of $M_R$ elements (blue) vs. normal distribution (red). (D) Evolution of $\lambda_{J(t)}$ during pre-training (top row), training (middle row), and testing (bottom row) with respect to the stability region (dash red line). Zoom-in plot of the right boundary of the stability region is shown below. $g=1.5$ is used in all plots.}
    \label{fig:R-FORCEDistribution}
\end{figure*}

We investigate the properties of $M_R$ (R-FORCE) by comparing it with the standard initialization $M$ (used in conventional FORCE) for a fixed scaling factor $g=1.5$. The spectral distribution of $M_R$ is shown in Fig.~\ref{fig:R-FORCEDistribution}~A. Fig.~\ref{fig:R-FORCEDistribution}~B presents a histogram of the entries in  $M_R$, revealing that most values follow a normal distribution with mean zero and standard deviation approximate $\pm 0.1 g$, similar to $M$, though a small number of outliers exist. To quantify how closely the distribution of $M_R$ matches that of $M$, we include a Quantile-Quantile plot in Fig.~\ref{fig:R-FORCEDistribution}~C. Indeed, the upper segment of the blue line represents the majority of entries in $M_R$ and is closely aligned with the red line that represents $M$. However, the lower segment of the blue line appears to depart from $M$ (quantiles $<-3$, which corresponds to elements $<-0.2g$) and indicates a small number of negative elements in $M_R$ that do not follow the normal distribution. These elements are shown in the zoom-in inset of Fig.~\ref{fig:R-FORCEDistribution}~B. Such tail distribution can be attributed to the constraints we applied to the generation of $\lambda_{M_R}$.

We investigate how the $M_R$ spectrum evolves during training and what properties contribute to the stability. In Fig.~\ref{fig:R-FORCEDistribution}~D, we follow the same setup used in the second row of Fig.~\ref{fig:FORCE_Results}~A. We plot $\lambda_{J_R(t)}$ along with the unit circle (solid red line) and the right boundary of the stability region (dashed red line) for representative time steps during different evolution stages: (i) pre-training, (ii) training and (iii) testing. We also zoom in on the right part of the spectrum for these three scenarios. Before training, $\lambda_J$ exhibits only minor fluctuations. This is expected since all the weights are fixed, and the fluctuations are due to the evolution of the $\bm{x(t)}$ term in Eq.~\ref{eq:Jacobian}. The zoom-in plots show that the fluctuations are not visually noticeable near the right component of the stability circle. Notably, this spectrum stays near the boundary for all time, which is the optimal condition for training and implies this network is trainable. During training, however, the spectrum undergoes significant changes, especially in its right component (scenario (ii)). We observe that the eigenvalues near the circle's right boundary oscillate back and forth. This behavior hints at the ability of the network to learn to generate periodic and time-evolving sequences through oscillations of specific eigenvalues across the critical circle boundary. We also observe such oscillations during testing (scenario (iii)). This indicates that the network learns and memorizes this evolution pattern during training and can reproduce it during testing.

\begin{figure*}[t]
    \centering
    \includegraphics[width=1\linewidth]{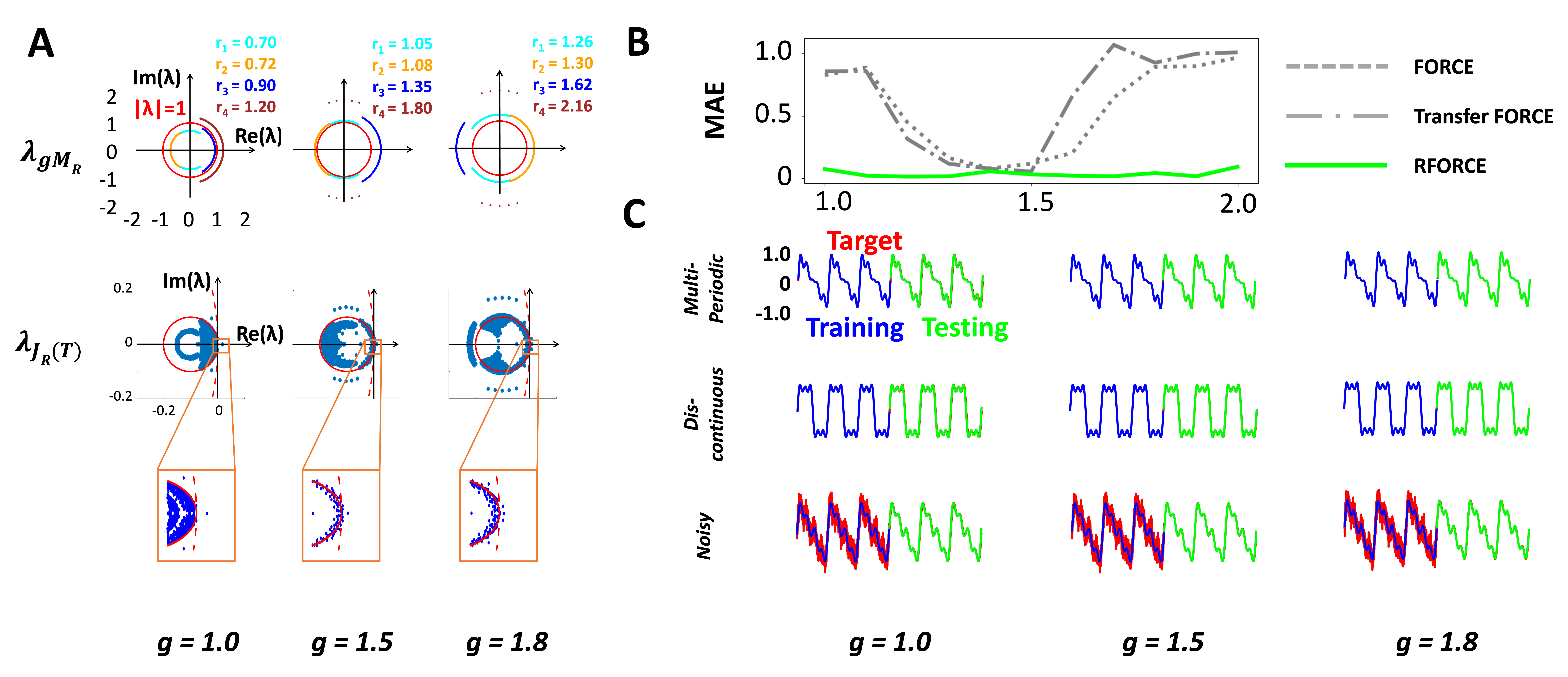}
    \caption{\textbf{R-FORCE with Rank1-RRNN.} (A) Spectrum of scaled connectivity matrix $gM_{R}$ (top), extended scaled connectivity matrix $g\Tilde{M}_{R}$ (mid), and Jacobian matrix $J_R(T)$ (bottom) for $g=1.0, 1.5, 1.8$. (B) Comparison of R-FORCE (green), and FORCE (dashed gray) error curves over the range of $g \in [1,2]$. (C) R-FORCE training (blue) and testing (green) performance for three examples of one-dimensional target functions (red). Compare with FORCE performance in Fig.~\ref{fig:FORCE_Results}. 
    }
    \label{fig:R-FORCE_eigen_distribution}
\end{figure*}

\subsection{R-FORCE Target-Learning}
\textbf{Rank1-RRNN Configuration:} We evaluate R-FORCE under varying degrees of chaos by applying it to Rank1-RRNNs, with $g=1.0, 1.5, 1.8$, as shown in Fig.~\ref{fig:R-FORCE_eigen_distribution}. The top row (Fig.~\ref{fig:R-FORCE_eigen_distribution}~A) shows the spectral distribution of $g M_R$, while the middle row displays the spectrum of the Jacobian $\lambda_{J(T)_R}$ at the end of training. We find that $\lambda_{J(T)_R}$ only has a few $J_R(T)$ eigenvalues outside the stability circle, and several positioned near the boundary of the stability region, which is consistent with the R-FORCE construction. Notably, while $\lambda_{J_R(T)}$ shows different non-trivial shapes for each $g$ value, the right component of the spectrum remains similar, demonstrating the contribution of the right component of the spectrum to stability.

\begin{figure}[t]
    \centering
    \includegraphics[width=1\linewidth]{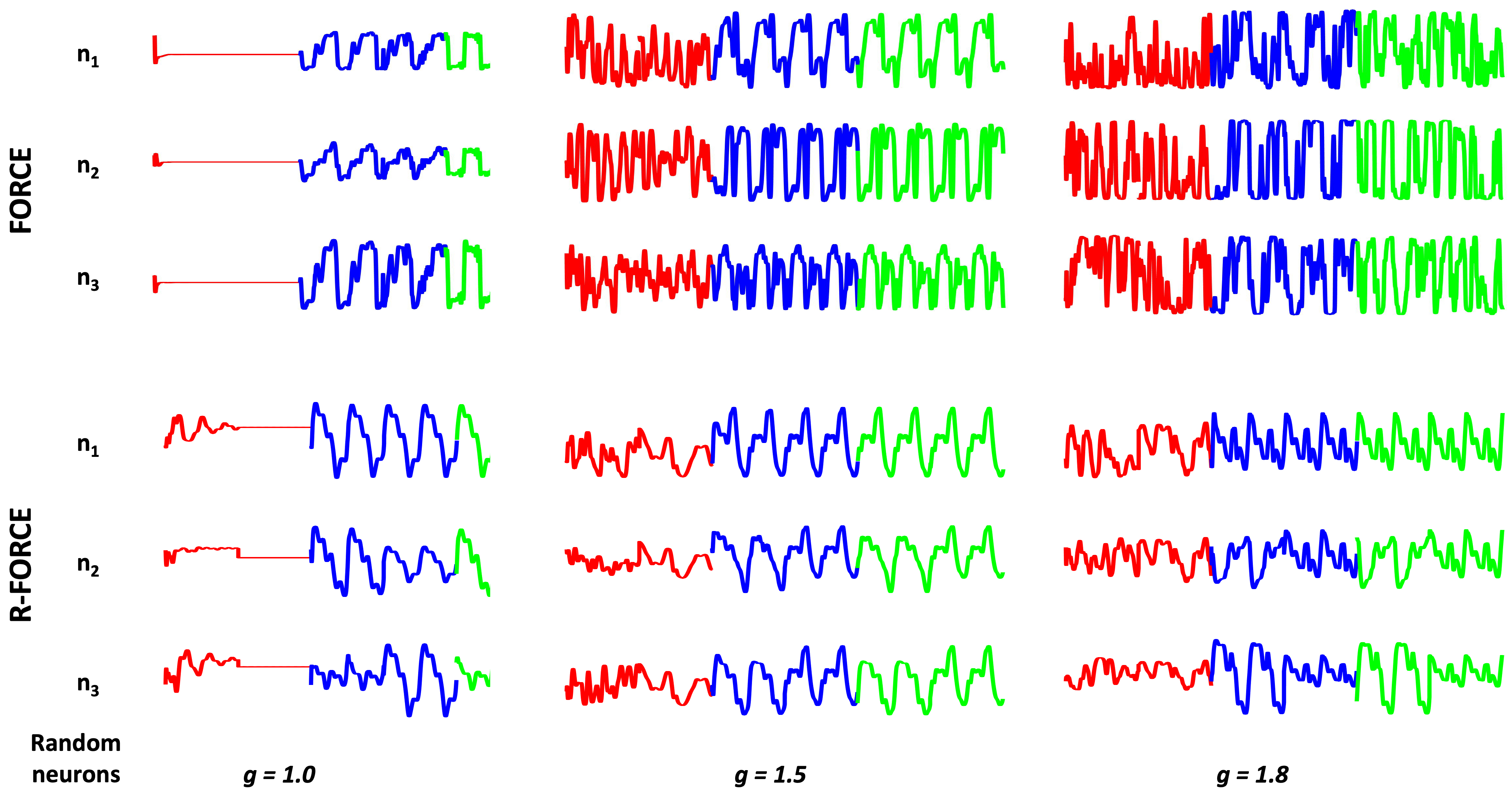}
    \caption{\textbf{Activity of Randomly Selected Neurons.} Activity of three neurons randomly selected from reservoirs of FORCE (top) and R-FORCE (bottom) Rank1-RRNN during pre-training (red), training (blue), and testing (green). The target is chosen as a superposition of four sine waves (multi-periodic). The evolution is shown for three representative values of $g$, $g=1.0, 1.5, 1.8$.}
    \label{fig:R-FORCE_neuronActivities}
\end{figure}

In Fig.~\ref{fig:R-FORCE_eigen_distribution}~B, We compare R-FORCE with FORCE and Transfer-FORCE~\cite{tamura2021transfer}, another alternative method, under the same setup as in Fig.~\ref{fig:FORCE_Results}~B. It shows that R-FORCE (solid green) achieves a broader operational regime, where the average error is controlled for all tested $g$ values, whereas FORCE (dashed gray) and Transfer-FORCE (dash-dot gray) fail to control the error for specific $g$ values. Additionally, when comparing the R-FORCE error with the lowest error that FORCE and Transfer-FORCE achieve at $g_{opt}$, we find that R-FORCE consistently outperforms them. The comparison is quantitatively summarized in Table~\ref{table:R-FORCE_FORCE_Comparison}, showcasing that R-FORCE outperforms FORCE and Transfer-FORCE in terms of MAE and $99\%$ CI under any $g$ values. On average, R-FORCE achieves an order of magnitude improvement over FORCE and Transfer-FORCE. Fig.~\ref{fig:R-FORCE_eigen_distribution}~C visually shows the performance of R-FORCE with three example target function examples: (i) multi-periodic, (ii) discontinuous, and (iii) noisy under three representative $g$ values ($g= 1.0, 1.5, 1.8$). It shows that R-FORCE generates visually indistinguishable outputs from the ground-truth target, even when training with a noisy target. For comparison, FORCE learning on these targets achieves the same output shape only for $g_{opt}$, and training with noise dampens the performance.

For further comparison between R-FORCE and FORCE, we randomly select three neurons from their networks with $g = 1.0, 1.5, 1.8$, and compare their activity evolution during pre-training, training, and testing, shown in Fig.~\ref{fig:R-FORCE_neuronActivities}. Before training, both R-FORCE and FORCE neuron activities converge to a constant, except for the convergence speed (R-FORCE appears to keep the information longer). However, when $g = 1.8$, significant differences emerge between the two networks during training. R-FORCE units approximately follow the shape (or part of it) and periods of the target, while FORCE units exhibit noise and a variety of periods and oscillation patterns. This behavior persists in R-FORCE units during testing, while FORCE units become noisier for non-optimal $g$ values. 

\begin{table*}[ht]
\centering
\caption{\textbf{Performance on Target-Learning Task with Rank1-RRNN.} MAE and $99\%$ CI ($\times 10^{-2}$) are computed for FORCE, Transfer-FORCE, and RFORCE for the multi-periodic target function with various values of $g$.}
\vspace{0.3cm}
\begin{tabular}{ c|c|c|c|c } 
\toprule
 & $g = 1.0$ & $g = 1.5$ & $g = 1.8$ & $g \in [1.0, 2.0]$ \\ 
 \midrule
Method & \multicolumn{4}{c}{(MAE $\pm$ CI) $\times 10^{-2}$} \\ 
\midrule
\midrule
FORCE & 82 $\pm$ 10 & 12 $\pm$ 11 & 89 $\pm$ 7.8 & 56 $\pm$ 11\\
\midrule
Transfer FORCE & 86 $\pm$ 8.6 & 5.5 $\pm$ 9.0 & 92 $\pm$ 29 & 63 $\pm$ 15\\
\midrule
\textbf{RFORCE (Ours)} & \textbf{7.5 $\pm$ 5.3} & \textbf{3.3 $\pm$ 1.9} & \textbf{4.4 $\pm$ 5.4} & \textbf{3.8 $\pm$ 1.6} \\
\bottomrule
\end{tabular}
\label{table:R-FORCE_FORCE_Comparison}
\end{table*}

\begin{figure}[t]
    \centering
    \includegraphics[width=1\linewidth]{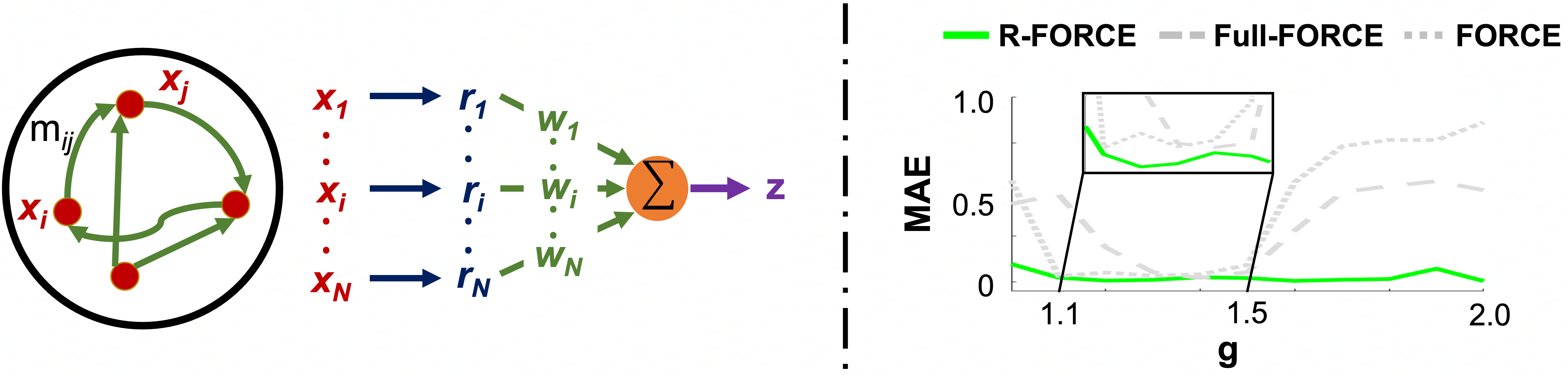}
    \caption{\textbf{R-FORCE Performance on RankN-RRNN.} (Left) RankN-RRNN architecture: The connectivity matrix (green) and the read-out connections (green) are optimized during training. (Right) Comparison of testing error curves for R-FORCE (green), FORCE (dashed gray) and Full-FORCE (dotted-gray) for an interval of $g \in [1,2]$.}
    \label{fig:otherArchitecture}
\end{figure}

\textbf{Higher-Rank RRNN Configuration}. We extend our analysis to the RankN-RRNN configuration, where the feedback loop is internal to the reservoir rather than through readout weights~\cite{sussillo2009generating}. This architecture more closely resembles standard RNNs. Fig.~\ref{fig:otherArchitecture}~(Left) illustrates the network topology.

We compare the performance achieved by R-FORCE (solid green) with FORCE (dashed gray) and Full-FORCE (dotted gray) in Fig.~\ref{fig:otherArchitecture}~(Right). Similar to the Rank1-RRNN result, R-FORCE achieves effective testing errors for all $g$ values in the RankN-RRNN setting, while the FORCE method is effective only in a narrow interval of $g \in [1.3,1.5]$. Full-FORCE achieves a wider effective interval, $g \in [1.1,1.5]$, but it is still inferior to R-FORCE. In particular, R-FORCE achieves a lower error and tighter CI than FORCE and Full-FORCE at their optimal $g_{opt}$ values ($g_{opt}=1.5$), as demonstrated in the zoom-in inset in Fig.~\ref{fig:otherArchitecture} and Table~\ref{table:R-FORCE_FORCE_Comparison_Other_Architecture}. Notably, R-FORCE with RankN-RRNN architecture achieves a better performance than its Rank1-RRNN counterpart under the optimal $g$ value ($g_{opt}=1.5$) and average circumstances ($g \in [1.0, 2.0]$).

\begin{table*}[ht]
\centering
\caption{\textbf{.} MAE and $99\%$ CI ($\times 10^{-2}$) are computed for FORCE, Full-FORCE, and RFORCE for the multi-periodic target function with various values of $g$.}
\vspace{0.3cm}
\begin{tabular}{c|c|c|c|c } 
\toprule
 & $g = 1.0$ & $g = 1.5$ & $g = 1.8$ & $g \in [1.0, 2.0]$ \\ 
 \midrule
Method & \multicolumn{4}{c}{(MAE $\pm$ CI) $\times 10^{-2}$} \\ 
\midrule
\midrule
Full-FORCE & 55 $\pm$ 46 & 9.2 $\pm$ 15 & 77 $\pm$ 5.9 & 41 $\pm$ 8.8\\
\midrule
FORCE & 42 $\pm$ 5.6 & 5.3 $\pm$ 9.3 & 52 $\pm$ 6.0 & 32 $\pm$ 6.2\\
\midrule
\textbf{RFORCE (Ours)} & \textbf{9.7 $\pm$ 12} & \textbf{2.1 $\pm$ 1.6} & \textbf{1.6 $\pm$ 0.9} & \textbf{2.7 $\pm$ 1.5} \\
\bottomrule
\end{tabular}

\label{table:R-FORCE_FORCE_Comparison_Other_Architecture}
\end{table*}

\section{Results}

\subsection{Multi-dimensional Targets Modeling}
Many real-world signals, such as human body joint movements, are inherently multi-dimensional. In this section, we demonstrate that R-FORCE can reproduce such complex signals and outperform other SOTA methods. Our experiments are conducted on the Physical Rehabilitation Movements dataset (UI-PRMD)~\cite{vakanski2018data}, which captures ten common physical therapy and rehabilitation exercises (e.g., deep squat, hurdle step). Each movement is represented by $22$ body joints in 3D space, resulting in a 66-dimensional time series. Joint locations are sampled at $30$ Hz.

\begin{figure}[ht]
    \centering
    \includegraphics[width=0.9\linewidth]{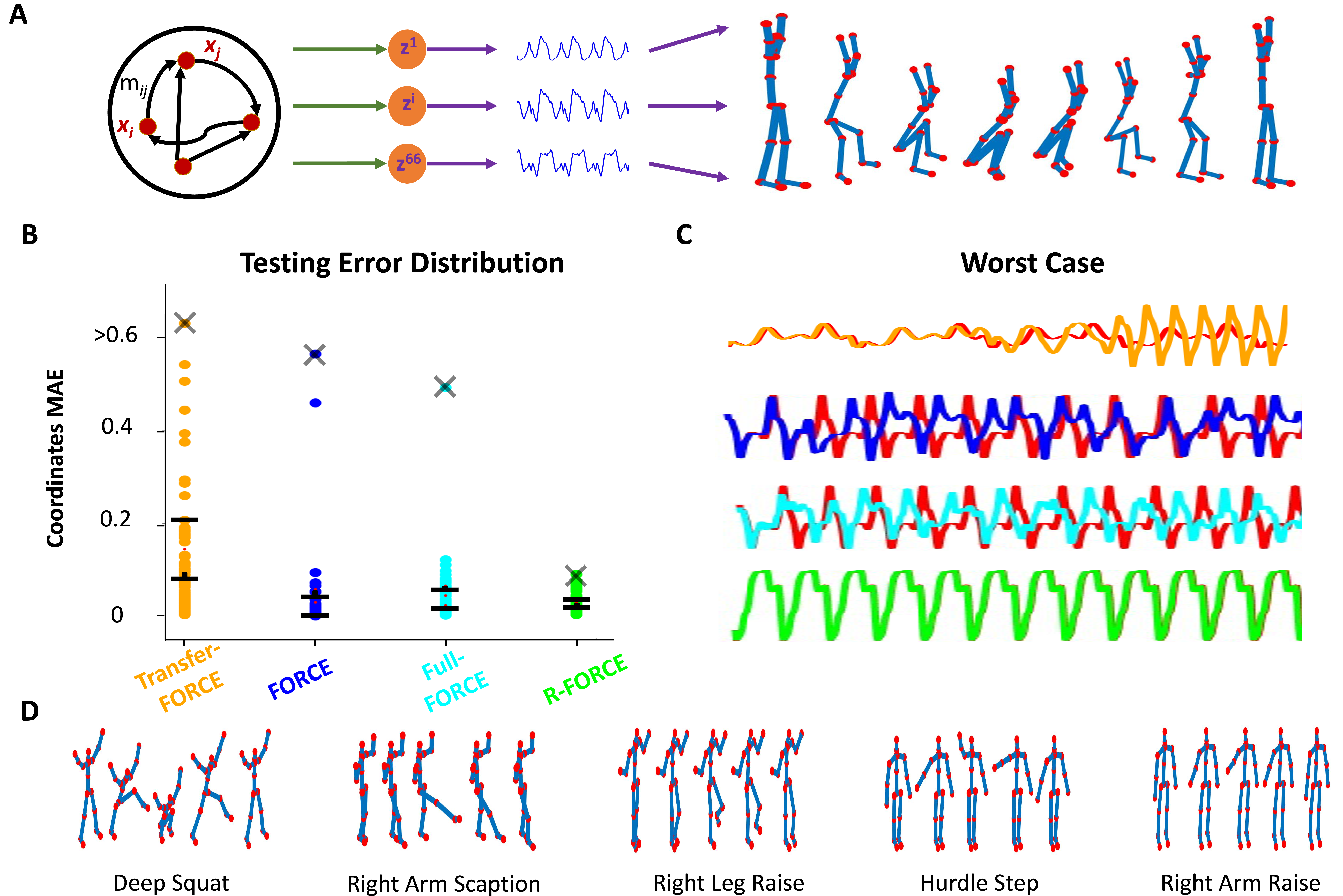}
    \caption{\textbf{Generation of body joints movements with Rank1-RRNN.} (A) Rank1-RRNN with 66 read-out units for the generation of joint movements time series. (B) Errors of Transfer-FORCE (orange), FORCE (blue), Full-FORCE (cyan), and R-FORCE (green) in generating the evolution of the deep squat movement. Each dot corresponds to a read-out unit result. The red cross stands for the read-out unit with the largest error (worst case) in each method. (C) The evolution of the read-out unit in the worst case with the ground-truth target (red). (D) Skeleton visualizations of the movements used for performance estimation.}
    \label{fig:R-FORCEBodyMovement}
\end{figure}

In our experiment, we employ Rank1-RRNN of two different sizes (Rank1-RRNN-400 and Rank1-RRNN-1000). To generate the movements, the network learns the coordinates of each joint separately, resulting in $66$ read-out units. As shown in Fig.~\ref{fig:R-FORCEBodyMovement}~A, the same reservoir is shared among read-out units. Each read-out unit also transfers information back to the reservoir through a feedback layer of Rank1-RRNN, which is not shown in the figure for simplicity. Because the number of frames in each movement is insufficient and coordinate data for each joint have different scales, we preprocess the data via augmentation and normalization. Specifically, we repeat the data $15$ times and concatenate them together. We then normalize the concatenated data to the range $[-1,1]$. Note that the generated sequences need to be scaled back to the original range during testing.

To visualize the model's behavior, we use the deep squat movement and Rank1-RRNN-1000 as a representative case. As shown in Fig.~\ref{fig:R-FORCEBodyMovement}~A, the target sequences exhibit highly structured and dynamic patterns. Fig.~\ref{fig:R-FORCEBodyMovement}~B shows the generation error for each coordinate, comparing R-FORCE (green) against Full-FORCE (cyan), FORCE (blue), and Transfer-FORCE (orange) at the optimal scaling value ($g = 1.5$). Each dot represents one coordinate sequence, while crosses mark the worst case per method. R-FORCE achieves the lowest overall error and tightest confidence interval (CI), indicating superior robustness. To further assess robustness, we present the worst-case generated sequences for each method in Fig.~\ref{fig:R-FORCEBodyMovement}~C. While Transfer-FORCE, FORCE, and Full-FORCE rapidly diverge from the ground truth (red) within a single episode, R-FORCE closely follows the target for a significantly longer duration. This highlights R-FORCE’s resilience, even under near-optimal conditions, where small deviations (outliers) can critically affect overall multi-dimensional generation. R-FORCE minimizes both CI and average error, delivering more reliable output.

\begin{table}[ht]
\tiny
\centering
\caption{\textbf{Performance on Target-Learning of Body Joints Movements with Rank1-RRNN.} FORCE, Transfer-FORCE, Full-Force, and RFORCE accuracy is evaluated for five different movements, $400$ and $1000$ neurons in the reservoir, and three values of the parameter $g$.}
\vspace{0.3cm}
\begin{tabular}{c|c|c|c|c|c} 
\toprule
 & Deep Squat & Arm Scaption & Leg Raise & Hurdle Step & Arm Raise \\ 
\midrule
Method & \multicolumn{5}{c}{(MAE $\pm$ CI) $\times 10^{-2}$} \\ 
\midrule\midrule                                 
 & \multicolumn{5}{c}{Optimal Regime ($g = 1.5$)} \\ 
\midrule
Transfer-FORCE-400 & 27 $\pm$ 7.6 & 34 $\pm$ 10 & 27 $\pm$ 10 & 29 $\pm$ 11 & 29 $\pm$ 8.3 \\ 
\midrule
Transfer-FORCE-1000 & 14 $\pm$ 6.4 & 19 $\pm$ 15 & 9.2 $\pm$ 7.0 & 14 $\pm$ 11 & 9.6 $\pm$ 8.6 \\ 
\midrule
Full-FORCE-400 & 8.5 $\pm$ 3.2 & 16 $\pm$ 6.4 & 5.7 $\pm$1.6 & 14 $\pm$ 6.8 & 11 $\pm$ 5.0 \\ 
\midrule
FORCE-400 & 6.9 $\pm$ 2.6 & 12 $\pm$ 5.8 & 4.0 $\pm$ 1.9 & 3.5 $\pm$ 2.4 & 5.0 $\pm$ 2.7 \\ 
\midrule
Full-FORCE-1000 & 4.4 $\pm$ 2.1 & 4.5 $\pm$ 2.3 & 3.0 $\pm$ 0.8 & 3.8 $\pm$ 1.0 & 4.0 $\pm$ 0.9 \\ 
\midrule
FORCE-1000 & 2.9 $\pm$ 2.9 & 5.2 $\pm$ 3.7 & 2.1 $\pm$ 1.2 & 3.8 $\pm$ 2.7 & 2.1 $\pm$ 1.5 \\ 
\midrule
\textbf{RFORCE-400 (Ours)} & \textbf{3.3 $\pm$ 1.4} & \textbf{4.1 $\pm$ 1.7} & \textbf{1.9 $\pm$ 0.7} & \textbf{3.3 $\pm$ 1.7} & \textbf{3.8 $\pm$ 2.3} \\ 
\midrule
\textbf{RFORCE-1000 (Ours)} & \textbf{1.0 $\pm$ 0.3} & \textbf{1.9 $\pm$ 0.9} & \textbf{0.9 $\pm$ 0.8} & \textbf{1.9 $\pm$ 4.9} & \textbf{1.5 $\pm$ 1.2} \\ 
\midrule
 & \multicolumn{5}{c}{Non-optimal (Quiescent) Regime ($g = 1.0$)} \\ 
\midrule\midrule
Transfer-FORCE-1000 & 69 $\pm$ 9.3 & 58 $\pm$ 9.7 & 32 $\pm$ 8.0 & 57 $\pm$ 26 & 48 $\pm$ 8.1 \\ 
\midrule
Full-FORCE-1000 & 88 $\pm$ 8.7 & 40 $\pm$ 11 & 78 $\pm$ 7.4 & 76 $\pm$ 17 & 82 $\pm$ 5.5 \\ 
\midrule
FORCE-100 & 81 $\pm$ 4.7 & 73 $\pm$ 4.5 & 63 $\pm$ 8.2 & 80 $\pm$ 6.5 & 72 $\pm$ 4.2 \\ 
\midrule
\textbf{\textbf{RFORCE-1000 (Ours)}} & \textbf{2.4 $\pm$ 0.6} & \textbf{2.6 $\pm$ 1.6} & \textbf{1.3 $\pm$ 0.3} & \textbf{5.7 $\pm$ 4.3} & \textbf{2.2 $\pm$ 1.0} \\ 
\midrule\midrule
 & \multicolumn{5}{c}{Non-optimal (Chaotic) Regime ($g = 1.8$)} \\ 
 
\midrule
Transfer-FORCE-1000 & 109 $\pm$ 17 & 99 $\pm$ 14 & 116 $\pm$ 16 & 104 $\pm$ 15 & 102 $\pm$ 13 \\ 
\midrule
Full-FORCE-1000 & 86 $\pm$ 29 & 75 $\pm$ 14 & 47 $\pm$ 7.5 & 76 $\pm$ 17 & 72 $\pm$ 13 \\ 
\midrule
FORCE-1000 & 63 $\pm$ 9.5 & 49 $\pm$ 11 & 21 $\pm$ 7.7 & 42 $\pm$ 9.0 & 42 $\pm$ 10 \\ 
\midrule
\textbf{\textbf{RFORCE-1000 (Ours)}} & \textbf{7.9 $\pm$ 3.9} & \textbf{4.5 $\pm$ 2.2} & \textbf{2.9 $\pm$ 2.1} & \textbf{8.7 $\pm$ 5.8} & \textbf{4.4 $\pm$ 2.0} \\
\bottomrule
\end{tabular}
\label{table:Body_Movement_Optimal}
\end{table}

We extend our evaluation across five different movements (examples shown in
Fig.~\ref{fig:R-FORCEBodyMovement}~D, 
under both Rank1-RRNN-400 and Rank1-RRNN-1000 architectures and three scaling values ($g = 1.0, 1.5, 1.8$). 
The quantitative comparison results between R-FORCE and other methods are included in Table~\ref{table:Body_Movement_Optimal}. With the same setup, R-FORCE consistently achieves the lowest error in all movements we have tested. In the cases of non-optimal choice of $g$ values, i.e., $g = 1.0, 1.8$, R-FORCE improves the performance by approximately one order of magnitude compared with other methods. For an optimal choice of $g$ value, i.e., $g = 1.5$, the R-FORCE with Rank1-RRNN-1000 outperforms all other network configurations and training methods. Furthermore, we observe that the application of R-FORCE to a smaller network, Rank1-RRNN-400, outperforms other methods using Rank1-RRNN-400 on all tested movements. It also performs similarly or better than other methods with Rank1-RRNN-1000. Thereby, R-FORCE-400 training can be used as a means for speed-up training (e.g., for the deep squat movement, R-FORCE-400 training takes about $8$min, while FORCE-1000 and Full-FORCE-1000 running times are approximately $50$min and $90$min, respectively).

\subsection{Spiking Neural Networks Modeling}
Although firing rate models are widely adopted for their simplicity, spiking neural networks (SNNs) offer a more biologically realistic representation of neuronal dynamics. To demonstrate the versatility of R-FORCE, we extend its application to spiking neural networks and examine its performance on two well-known spiking neuron models, namely Theta~\cite{ermentrout2008ermentrout} and Izhikevich~\cite{izhikevich2003simple} models, in line with their FORCE counterparts implemented in~\cite{nicola2017supervised}.

Before delving into the experiment, we briefly describe the Theta and Izhikevich Neurons. Each neuron in the network follows these evolution equations: \\
Theta model:
\begin{align}
    \dot{\theta_i} = (1 - cos(\theta_i)) + \pi^2 (1 + cos(\theta_i))(I).
\end{align}
Izhikevich model:
\begin{align}
        C \dot{v_i} & = k(v_i - v_r)(v_i - v_t) - u_i + I,\\
        \dot{u_i} &= a(b(v_i - v_r) - u_i).
\end{align}
Here, $I$ is current, $\theta_i$ and $v_i$ are voltage variables, and $u_i$ is adaption current. $C$ is membrane capacitance, $v_r$ and $v_t$ are resting membrane potential and membrane threshold potential, respectively. $a$ is the reciprocal of the adaptation time constant, $k$ is the control action potential half-width, and $b$ is the control resonance properties of the model.

\begin{figure}[ht]
    \centering
    \includegraphics[width=0.9\linewidth]{figures/Figure9_birdsong_spec.png}
    \caption{\textbf{Spiking neural network with Rank1-RRNN.} (A-B) Generation of a sine wave with Theta model: (A) 1Hz sine wave. (B) Voltage trace of three randomly selected Theta neurons. (C-G) Generation of birdsong spectrum with Izhikevich neurons: (C) Sound wave sample of birdsong. (D). Voltage trace of three randomly selected neurons. (E) MAE of FORCE (blue) and R-FORCE(green) in generating the spectrum of birdsong. Each dot corresponds to a read-out unit, and black bars are $99$\% confidence interval. (F) Comparison between FORCE (blue) and R-FORCE(green) of a randomly selected band over flat and peak regions. (G) Spectrum comparison among FORCE (top row), R-FORCE(middle row) prediction, and ground truth (bottom row). The right column is the zoom-in plot of the selected area (red rectangle) in the Left column.}
    \label{fig:spiking_neural_network}
\end{figure}

We first apply R-FORCE to a simple $1$ Hz sine wave target, shown in Fig.~\ref{fig:spiking_neural_network}~A. Fig.~\ref{fig:spiking_neural_network}~B depicts the training process for three randomly selected neurons, where pre-training (red) shows sparse voltage traces without specific patterns. Each neuron learns its pattern during training (blue), which is followed during testing (green). We perform five repetitions for each period (pre-training, training, and testing). The results demonstrate that R-FORCE achieves less error ($0.122$) than FORCE ($0.193$) in this sine wave task.

To assess performance on more complex signals, we employ the Izhikevich model to reproduce the spectrogram of birdsong, rather than the raw audio waveform, following the approach in ~\cite{nicola2017supervised}. A sample of the raw birdsong audio is shown in Fig.~\ref{fig:spiking_neural_network}~C, while its 33-band spectrogram (used as the target) appears in the third row of Fig.~\ref{fig:spiking_neural_network}~G. Following the setup in ~\cite{crandall2014neural}, we perform $5$s (1 repetition) of idling network, $85$s (17 repetitions) of training, and $10$s (2 repetitions) of testing. Fig.~\ref{fig:spiking_neural_network}~D displays voltage traces of three randomly selected neurons, with each trace consisting of $1$ repetition of pre-training (red), four repetitions of training (blue), and one repetition of testing (green). We can observe from this that during training and testing, each neuron learns and follows its pattern. In Fig.~\ref{fig:spiking_neural_network}~EFG, we compare the performance of FORCE and R-FORCE in the birdsong task. Fig.~\ref{fig:spiking_neural_network}~E shows the error distributions of FORCE and R-FORCE over $33$ bands, where each dot represents a band, and the black bar indicates the $99\%$ CI. Our findings reveal that R-FORCE 
(
$
0.0254 \pm 0.0055$
)
obtains a lower average error and tighter CI than FORCE($
0.0308 \pm 0.0063$
). In Fig.~\ref{fig:spiking_neural_network}~F, we showcase two common forms in the targets: flat regions with idling signals and peak regions with pulse signals. We select one example for each region from a randomly chosen target band (red). The results show that R-FORCE (green) not only overlaps better with the target than FORCE (blue) in peak regions but also exhibits less noise in flat regions. Fig.~\ref{fig:spiking_neural_network}~G displays spectrograms reproduced by FORCE and R-FORCE. The difference between FORCE, R-FORCE, and the ground truth is almost negligible in the left column. However, in the right column of Fig.~\ref{fig:spiking_neural_network}~G (red region), the ground truth is clean, and R-FORCE exhibits less noise than FORCE. This observation confirms that R-FORCE performs better in flat regions.

\section{Discussion}

Various training approaches have been proposed for RRNNs to tackle the target-learning task where the network is tasked to generate a predefined sequence without guiding input. Among them, FORCE learning achieves success across different targets. However, it fails to generalize to different network dynamics and chaos, which is determined by a scaling parameter $g$. We introduce the R-FORCE initialization for the RRNN connectivity matrix to increase its robustness to chaos. This new initialization follows four generating principles that constrain the spectrum of the network Jacobian matrix to remain in a well-balanced regime conducive to stability and successful training.

Our analysis of the R-FORCE distribution indicates that it is similar to the normal distribution with several exceptions. For example, a close inspection of the R-FORCE distribution shows negative weights that deviate from the normal distribution and form a longer tail on the negative axis. While RRNNs wired from a normal distribution were operational exclusively in the parameter range of edge-of-chaos, those negative weights in the R-FORCE distribution succeed in confining the spectrum of the Jacobian under different chaos. Such a property suggests that negative (inhibitory) weights are essential in balancing the network dynamics and could be configured with the initialization we infer. Indeed, in biological neural networks, the balance of excitation and inhibition is a critical requirement for proper network function~\cite{shadlen1998variable,haider2006neocortical,rubin2017balanced}. Our work strengthens these findings and provides additional insights into the generating rules of the distribution of connectivity that generic networks need to exhibit robust functionality.  

R-FORCE initialization places the network in an optimal configuration beforehand, making it well-suited for stable training and robust performance. Notably, R-FORCE can even stabilize the training and functionality of Rank1-RRNN, where only a fraction of the weights are optimized during training. This highlights the importance of initialization in determining network performance. Our findings are potentially indicative of how generic networks of neurons are optimally structured to be `amenable for learning' and potentially an interface between computational research of optimal wiring of RRNN and anatomical wiring of brain networks (i.e., connectomics ~\cite{sporns2012discovering,cook2019whole,bates2020complete}). 

We demonstrate that R-FORCE outperforms FORCE learning and other SOTA methods by achieving a smaller average error and tighter confidence interval, minimizing the occurrence of outliers where the generated sequence deviates from the target in a short time. This is particularly critical in modeling multi-dimensional targets, where the likelihood of outliers increases with dimensionality. We demonstrate the effectiveness of R-FORCE in modeling five 3D body joint movements. Each movement is composed of 66 sequences with intricate patterns. Our experiments indicate that R-FORCE can generate these sequences with a typical total error of $\mathcal{O}(10^{-2})$ and not exceeding $\mathcal{O}(10^{-1})$. Furthermore, we show that R-FORCE achieves SOTA performance on spiking neural networks, such as the Theta and Izhikevich models. R-FORCE obtains $30\%$ less error than FORCE for the Theta model performing sine wave. Furthermore, we reproduce a birdsong spectrogram with $33$ bands via the Izhikevich model. The results show that R-FORCE achieves about $20\%$ less average error and $15\%$ smaller CI than FORCE.

In summary, R-FORCE initialization facilitates a robust performance of the target-learning task. We have demonstrated the principles upon which R-FORCE distribution is generated and provided an algorithmic way to generate the distribution. Our results are generic and applicable for a wide range of networks and tasks since we demonstrate the effectiveness of R-FORCE on challenging targets (e.g., noisy and highly dynamic), network architectures (e.g., low-rank RRNN), and neuron forms (e.g., spiking neural network).

\section{References}
\bibliographystyle{elsarticle-num-names}
\bibliography{rforce_ref}

\begin{thebibliography}{44}
\expandafter\ifx\csname natexlab\endcsname\relax\def\natexlab#1{#1}\fi
\providecommand{\url}[1]{\texttt{#1}}
\providecommand{\href}[2]{#2}
\providecommand{\path}[1]{#1}
\providecommand{\DOIprefix}{doi:}
\providecommand{\ArXivprefix}{arXiv:}
\providecommand{\URLprefix}{URL: }
\providecommand{\Pubmedprefix}{pmid:}
\providecommand{\doi}[1]{\href{http://dx.doi.org/#1}{\path{#1}}}
\providecommand{\Pubmed}[1]{\href{pmid:#1}{\path{#1}}}
\providecommand{\bibinfo}[2]{#2}
\ifx\xfnm\relax \def\xfnm[#1]{\unskip,\space#1}\fi
\bibitem[{Yao et~al.(2014)Yao, Peng, Zhang, Yu, Zweig, and Shi}]{yao2014spoken}
\bibinfo{author}{K.~Yao}, \bibinfo{author}{B.~Peng},
  \bibinfo{author}{Y.~Zhang}, \bibinfo{author}{D.~Yu},
  \bibinfo{author}{G.~Zweig}, \bibinfo{author}{Y.~Shi},
\newblock \bibinfo{title}{Spoken language understanding using long short-term
  memory neural networks},
\newblock in: \bibinfo{booktitle}{2014 IEEE Spoken Language Technology Workshop
  (SLT)}, \bibinfo{organization}{IEEE}, \bibinfo{year}{2014}, pp.
  \bibinfo{pages}{189--194}.
\bibitem[{Graves et~al.(2013)Graves, Mohamed, and Hinton}]{graves2013speech}
\bibinfo{author}{A.~Graves}, \bibinfo{author}{A.-r. Mohamed},
  \bibinfo{author}{G.~Hinton},
\newblock \bibinfo{title}{Speech recognition with deep recurrent neural
  networks},
\newblock in: \bibinfo{booktitle}{2013 IEEE international conference on
  acoustics, speech and signal processing}, \bibinfo{organization}{IEEE},
  \bibinfo{year}{2013}, pp. \bibinfo{pages}{6645--6649}.
\bibitem[{Eck and Schmidhuber(2002)}]{eck2002learning}
\bibinfo{author}{D.~Eck}, \bibinfo{author}{J.~Schmidhuber},
\newblock \bibinfo{title}{Learning the long-term structure of the blues},
\newblock in: \bibinfo{booktitle}{International Conference on Artificial Neural
  Networks}, \bibinfo{organization}{Springer}, \bibinfo{year}{2002}, pp.
  \bibinfo{pages}{284--289}.
\bibitem[{Shlizerman et~al.(2018)Shlizerman, Dery, Schoen, and
  Kemelmacher-Shlizerman}]{shlizerman2018audio}
\bibinfo{author}{E.~Shlizerman}, \bibinfo{author}{L.~Dery},
  \bibinfo{author}{H.~Schoen}, \bibinfo{author}{I.~Kemelmacher-Shlizerman},
\newblock \bibinfo{title}{Audio to body dynamics},
\newblock \bibinfo{journal}{IEEE Computer Society Conference on Computer Vision
  and Pattern Recognition (CVPR)}  (\bibinfo{year}{2018}).
\bibitem[{Baccouche et~al.(2011)Baccouche, Mamalet, Wolf, Garcia, and
  Baskurt}]{baccouche2011sequential}
\bibinfo{author}{M.~Baccouche}, \bibinfo{author}{F.~Mamalet},
  \bibinfo{author}{C.~Wolf}, \bibinfo{author}{C.~Garcia},
  \bibinfo{author}{A.~Baskurt},
\newblock \bibinfo{title}{Sequential deep learning for human action
  recognition},
\newblock in: \bibinfo{booktitle}{International workshop on human behavior
  understanding}, \bibinfo{organization}{Springer}, \bibinfo{year}{2011}, pp.
  \bibinfo{pages}{29--39}.
\bibitem[{Su and Shlizerman(2020)}]{su2020clustering}
\bibinfo{author}{K.~Su}, \bibinfo{author}{E.~Shlizerman},
\newblock \bibinfo{title}{Clustering and recognition of spatiotemporal features
  through interpretable embedding of sequence to sequence recurrent neural
  networks},
\newblock \bibinfo{journal}{Frontiers in artificial intelligence}
  \bibinfo{volume}{3} (\bibinfo{year}{2020}) \bibinfo{pages}{70}.
\bibitem[{Hochreiter and Schmidhuber(1997)}]{hochreiter1997long}
\bibinfo{author}{S.~Hochreiter}, \bibinfo{author}{J.~Schmidhuber},
\newblock \bibinfo{title}{Long short-term memory},
\newblock \bibinfo{journal}{Neural computation} \bibinfo{volume}{9}
  (\bibinfo{year}{1997}) \bibinfo{pages}{1735--1780}.
\bibitem[{Cho et~al.(2014)Cho, Van~Merri{\"e}nboer, Gulcehre, Bahdanau,
  Bougares, Schwenk, and Bengio}]{cho2014learning}
\bibinfo{author}{K.~Cho}, \bibinfo{author}{B.~Van~Merri{\"e}nboer},
  \bibinfo{author}{C.~Gulcehre}, \bibinfo{author}{D.~Bahdanau},
  \bibinfo{author}{F.~Bougares}, \bibinfo{author}{H.~Schwenk},
  \bibinfo{author}{Y.~Bengio},
\newblock \bibinfo{title}{Learning phrase representations using rnn
  encoder-decoder for statistical machine translation},
\newblock \bibinfo{journal}{arXiv preprint arXiv:1406.1078}
  (\bibinfo{year}{2014}).
\bibitem[{Jaeger(2001)}]{jaeger2001echo}
\bibinfo{author}{H.~Jaeger},
\newblock \bibinfo{title}{The “echo state” approach to analysing and
  training recurrent neural networks-with an erratum note},
\newblock \bibinfo{journal}{Bonn, Germany: German National Research Center for
  Information Technology GMD Technical Report} \bibinfo{volume}{148}
  (\bibinfo{year}{2001}) \bibinfo{pages}{13}.
\bibitem[{Buonomano and Merzenich(1995)}]{buonomano1995temporal}
\bibinfo{author}{D.~V. Buonomano}, \bibinfo{author}{M.~M. Merzenich},
\newblock \bibinfo{title}{Temporal information transformed into a spatial code
  by a neural network with realistic properties},
\newblock \bibinfo{journal}{Science} \bibinfo{volume}{267}
  (\bibinfo{year}{1995}) \bibinfo{pages}{1028--1030}.
\bibitem[{Maass et~al.(2002)Maass, Natschl{\"a}ger, and
  Markram}]{maass2002real}
\bibinfo{author}{W.~Maass}, \bibinfo{author}{T.~Natschl{\"a}ger},
  \bibinfo{author}{H.~Markram},
\newblock \bibinfo{title}{Real-time computing without stable states: A new
  framework for neural computation based on perturbations},
\newblock \bibinfo{journal}{Neural computation} \bibinfo{volume}{14}
  (\bibinfo{year}{2002}) \bibinfo{pages}{2531--2560}.
\bibitem[{Jaeger(2003)}]{jaeger2003adaptive}
\bibinfo{author}{H.~Jaeger},
\newblock \bibinfo{title}{Adaptive nonlinear system identification with echo
  state networks},
\newblock in: \bibinfo{booktitle}{Advances in neural information processing
  systems}, \bibinfo{year}{2003}, pp. \bibinfo{pages}{609--616}.
\bibitem[{Sussillo and Abbott(2009)}]{sussillo2009generating}
\bibinfo{author}{D.~Sussillo}, \bibinfo{author}{L.~F. Abbott},
\newblock \bibinfo{title}{Generating coherent patterns of activity from chaotic
  neural networks},
\newblock \bibinfo{journal}{Neuron} \bibinfo{volume}{63} (\bibinfo{year}{2009})
  \bibinfo{pages}{544--557}.
\bibitem[{Sompolinsky et~al.(1988)Sompolinsky, Crisanti, and
  Sommers}]{sompolinsky1988chaos}
\bibinfo{author}{H.~Sompolinsky}, \bibinfo{author}{A.~Crisanti},
  \bibinfo{author}{H.-J. Sommers},
\newblock \bibinfo{title}{Chaos in random neural networks},
\newblock \bibinfo{journal}{Physical review letters} \bibinfo{volume}{61}
  (\bibinfo{year}{1988}) \bibinfo{pages}{259}.
\bibitem[{DePasquale et~al.(2018)DePasquale, Cueva, Rajan, Abbott
  et~al.}]{depasquale2018full}
\bibinfo{author}{B.~DePasquale}, \bibinfo{author}{C.~J. Cueva},
  \bibinfo{author}{K.~Rajan}, \bibinfo{author}{L.~Abbott}, et~al.,
\newblock \bibinfo{title}{full-force: A target-based method for training
  recurrent networks},
\newblock \bibinfo{journal}{PloS one} \bibinfo{volume}{13}
  (\bibinfo{year}{2018}) \bibinfo{pages}{e0191527}.
\bibitem[{Hardt et~al.(2016)Hardt, Recht, and Singer}]{hardt2016train}
\bibinfo{author}{M.~Hardt}, \bibinfo{author}{B.~Recht},
  \bibinfo{author}{Y.~Singer},
\newblock \bibinfo{title}{Train faster, generalize better: Stability of
  stochastic gradient descent},
\newblock in: \bibinfo{booktitle}{International conference on machine
  learning}, \bibinfo{organization}{PMLR}, \bibinfo{year}{2016}, pp.
  \bibinfo{pages}{1225--1234}.
\bibitem[{Pascanu et~al.(2013)Pascanu, Mikolov, and
  Bengio}]{pascanu2013difficulty}
\bibinfo{author}{R.~Pascanu}, \bibinfo{author}{T.~Mikolov},
  \bibinfo{author}{Y.~Bengio},
\newblock \bibinfo{title}{On the difficulty of training recurrent neural
  networks},
\newblock in: \bibinfo{booktitle}{International conference on machine
  learning}, \bibinfo{year}{2013}, pp. \bibinfo{pages}{1310--1318}.
\bibitem[{Chen et~al.(2017)Chen, Badrinarayanan, Lee, and
  Rabinovich}]{chen2017gradnorm}
\bibinfo{author}{Z.~Chen}, \bibinfo{author}{V.~Badrinarayanan},
  \bibinfo{author}{C.-Y. Lee}, \bibinfo{author}{A.~Rabinovich},
\newblock \bibinfo{title}{Gradnorm: Gradient normalization for adaptive loss
  balancing in deep multitask networks},
\newblock \bibinfo{journal}{arXiv preprint arXiv:1711.02257}
  (\bibinfo{year}{2017}).
\bibitem[{Graves(2013)}]{graves2013generating}
\bibinfo{author}{A.~Graves},
\newblock \bibinfo{title}{Generating sequences with recurrent neural networks},
\newblock \bibinfo{journal}{arXiv preprint arXiv:1308.0850}
  (\bibinfo{year}{2013}).
\bibitem[{Mhammedi et~al.(2017)Mhammedi, Hellicar, Rahman, and
  Bailey}]{mhammedi2017efficient}
\bibinfo{author}{Z.~Mhammedi}, \bibinfo{author}{A.~Hellicar},
  \bibinfo{author}{A.~Rahman}, \bibinfo{author}{J.~Bailey},
\newblock \bibinfo{title}{Efficient orthogonal parametrisation of recurrent
  neural networks using householder reflections},
\newblock in: \bibinfo{booktitle}{Proceedings of the 34th International
  Conference on Machine Learning-Volume 70}, \bibinfo{organization}{JMLR. org},
  \bibinfo{year}{2017}, pp. \bibinfo{pages}{2401--2409}.
\bibitem[{Mikolov et~al.(2014)Mikolov, Joulin, Chopra, Mathieu, and
  Ranzato}]{mikolov2014learning}
\bibinfo{author}{T.~Mikolov}, \bibinfo{author}{A.~Joulin},
  \bibinfo{author}{S.~Chopra}, \bibinfo{author}{M.~Mathieu},
  \bibinfo{author}{M.~Ranzato},
\newblock \bibinfo{title}{Learning longer memory in recurrent neural networks},
\newblock \bibinfo{journal}{arXiv preprint arXiv:1412.7753}
  (\bibinfo{year}{2014}).
\bibitem[{Le et~al.(2015)Le, Jaitly, and Hinton}]{le2015simple}
\bibinfo{author}{Q.~V. Le}, \bibinfo{author}{N.~Jaitly}, \bibinfo{author}{G.~E.
  Hinton},
\newblock \bibinfo{title}{A simple way to initialize recurrent networks of
  rectified linear units},
\newblock \bibinfo{journal}{arXiv preprint arXiv:1504.00941}
  (\bibinfo{year}{2015}).
\bibitem[{Maduranga et~al.(2019)Maduranga, Helfrich, and
  Ye}]{maduranga2019complex}
\bibinfo{author}{K.~D. Maduranga}, \bibinfo{author}{K.~E. Helfrich},
  \bibinfo{author}{Q.~Ye},
\newblock \bibinfo{title}{Complex unitary recurrent neural networks using
  scaled cayley transform},
\newblock in: \bibinfo{booktitle}{Proceedings of the AAAI Conference on
  Artificial Intelligence}, volume~\bibinfo{volume}{33}, \bibinfo{year}{2019},
  pp. \bibinfo{pages}{4528--4535}.
\bibitem[{Wisdom et~al.(2016)Wisdom, Powers, Hershey, Le~Roux, and
  Atlas}]{wisdom2016full}
\bibinfo{author}{S.~Wisdom}, \bibinfo{author}{T.~Powers},
  \bibinfo{author}{J.~Hershey}, \bibinfo{author}{J.~Le~Roux},
  \bibinfo{author}{L.~Atlas},
\newblock \bibinfo{title}{Full-capacity unitary recurrent neural networks},
\newblock in: \bibinfo{booktitle}{Advances in neural information processing
  systems}, \bibinfo{year}{2016}, pp. \bibinfo{pages}{4880--4888}.
\bibitem[{Arjovsky et~al.(2016)Arjovsky, Shah, and
  Bengio}]{arjovsky2016unitary}
\bibinfo{author}{M.~Arjovsky}, \bibinfo{author}{A.~Shah},
  \bibinfo{author}{Y.~Bengio},
\newblock \bibinfo{title}{Unitary evolution recurrent neural networks},
\newblock in: \bibinfo{booktitle}{International Conference on Machine
  Learning}, \bibinfo{year}{2016}, pp. \bibinfo{pages}{1120--1128}.
\bibitem[{Helfrich et~al.(2017)Helfrich, Willmott, and
  Ye}]{helfrich2017orthogonal}
\bibinfo{author}{K.~Helfrich}, \bibinfo{author}{D.~Willmott},
  \bibinfo{author}{Q.~Ye},
\newblock \bibinfo{title}{Orthogonal recurrent neural networks with scaled
  cayley transform},
\newblock \bibinfo{journal}{arXiv preprint arXiv:1707.09520}
  (\bibinfo{year}{2017}).
\bibitem[{Vorontsov et~al.(2017)Vorontsov, Trabelsi, Kadoury, and
  Pal}]{vorontsov2017orthogonality}
\bibinfo{author}{E.~Vorontsov}, \bibinfo{author}{C.~Trabelsi},
  \bibinfo{author}{S.~Kadoury}, \bibinfo{author}{C.~Pal},
\newblock \bibinfo{title}{On orthogonality and learning recurrent networks with
  long term dependencies},
\newblock in: \bibinfo{booktitle}{Proceedings of the 34th International
  Conference on Machine Learning-Volume 70}, \bibinfo{organization}{JMLR. org},
  \bibinfo{year}{2017}, pp. \bibinfo{pages}{3570--3578}.
\bibitem[{Chang et~al.(2019)Chang, Chen, Haber, and
  Chi}]{chang2019antisymmetricrnn}
\bibinfo{author}{B.~Chang}, \bibinfo{author}{M.~Chen},
  \bibinfo{author}{E.~Haber}, \bibinfo{author}{E.~H. Chi},
\newblock \bibinfo{title}{Antisymmetricrnn: A dynamical system view on
  recurrent neural networks},
\newblock \bibinfo{journal}{arXiv preprint arXiv:1902.09689}
  (\bibinfo{year}{2019}).
\bibitem[{Haber and Ruthotto(2017)}]{haber2017stable}
\bibinfo{author}{E.~Haber}, \bibinfo{author}{L.~Ruthotto},
\newblock \bibinfo{title}{Stable architectures for deep neural networks},
\newblock \bibinfo{journal}{Inverse Problems} \bibinfo{volume}{34}
  (\bibinfo{year}{2017}) \bibinfo{pages}{014004}.
\bibitem[{Laurent and von Brecht(2016)}]{laurent2016recurrent}
\bibinfo{author}{T.~Laurent}, \bibinfo{author}{J.~von Brecht},
\newblock \bibinfo{title}{A recurrent neural network without chaos},
\newblock \bibinfo{journal}{arXiv preprint arXiv:1612.06212}
  (\bibinfo{year}{2016}).
\bibitem[{Vakanski et~al.(2018)Vakanski, Jun, Paul, and
  Baker}]{vakanski2018data}
\bibinfo{author}{A.~Vakanski}, \bibinfo{author}{H.-p. Jun},
  \bibinfo{author}{D.~Paul}, \bibinfo{author}{R.~Baker},
\newblock \bibinfo{title}{A data set of human body movements for physical
  rehabilitation exercises},
\newblock \bibinfo{journal}{Data} \bibinfo{volume}{3} (\bibinfo{year}{2018})
  \bibinfo{pages}{2}.
\bibitem[{Bunch et~al.(1978)Bunch, Nielsen, and Sorensen}]{bunch1978rank}
\bibinfo{author}{J.~R. Bunch}, \bibinfo{author}{C.~P. Nielsen},
  \bibinfo{author}{D.~C. Sorensen},
\newblock \bibinfo{title}{Rank-one modification of the symmetric eigenproblem},
\newblock \bibinfo{journal}{Numerische Mathematik} \bibinfo{volume}{31}
  (\bibinfo{year}{1978}) \bibinfo{pages}{31--48}.
\bibitem[{Van~Loan and Golub(1983)}]{van1983matrix}
\bibinfo{author}{C.~F. Van~Loan}, \bibinfo{author}{G.~H. Golub},
  \bibinfo{title}{Matrix computations}, \bibinfo{publisher}{Johns Hopkins
  University Press}, \bibinfo{year}{1983}.
\bibitem[{Tamura and Tanaka(2021)}]{tamura2021transfer}
\bibinfo{author}{H.~Tamura}, \bibinfo{author}{G.~Tanaka},
\newblock \bibinfo{title}{Transfer-rls method and transfer-force learning for
  simple and fast training of reservoir computing models},
\newblock \bibinfo{journal}{Neural Networks} \bibinfo{volume}{143}
  (\bibinfo{year}{2021}) \bibinfo{pages}{550--563}.
\bibitem[{Ermentrout(2008)}]{ermentrout2008ermentrout}
\bibinfo{author}{B.~Ermentrout},
\newblock \bibinfo{title}{Ermentrout-kopell canonical model},
\newblock \bibinfo{journal}{Scholarpedia} \bibinfo{volume}{3}
  (\bibinfo{year}{2008}) \bibinfo{pages}{1398}.
\bibitem[{Izhikevich(2003)}]{izhikevich2003simple}
\bibinfo{author}{E.~M. Izhikevich},
\newblock \bibinfo{title}{Simple model of spiking neurons},
\newblock \bibinfo{journal}{IEEE Transactions on neural networks}
  \bibinfo{volume}{14} (\bibinfo{year}{2003}) \bibinfo{pages}{1569--1572}.
\bibitem[{Nicola and Clopath(2017)}]{nicola2017supervised}
\bibinfo{author}{W.~Nicola}, \bibinfo{author}{C.~Clopath},
\newblock \bibinfo{title}{Supervised learning in spiking neural networks with
  force training},
\newblock \bibinfo{journal}{Nature communications} \bibinfo{volume}{8}
  (\bibinfo{year}{2017}) \bibinfo{pages}{1--15}.
\bibitem[{Crandall and Nick(2014)}]{crandall2014neural}
\bibinfo{author}{S.~Crandall}, \bibinfo{author}{T.~Nick},
\newblock \bibinfo{title}{Neural population spiking activity during singing:
  adult and longitudinal developmental recordings in the zebra finch},
\newblock \bibinfo{journal}{CRCNS. org http://dx. doi. org/10.6080/K0NP22C8}
  (\bibinfo{year}{2014}).
\bibitem[{Shadlen and Newsome(1998)}]{shadlen1998variable}
\bibinfo{author}{M.~N. Shadlen}, \bibinfo{author}{W.~T. Newsome},
\newblock \bibinfo{title}{The variable discharge of cortical neurons:
  implications for connectivity, computation, and information coding},
\newblock \bibinfo{journal}{Journal of neuroscience} \bibinfo{volume}{18}
  (\bibinfo{year}{1998}) \bibinfo{pages}{3870--3896}.
\bibitem[{Haider et~al.(2006)Haider, Duque, Hasenstaub, and
  McCormick}]{haider2006neocortical}
\bibinfo{author}{B.~Haider}, \bibinfo{author}{A.~Duque}, \bibinfo{author}{A.~R.
  Hasenstaub}, \bibinfo{author}{D.~A. McCormick},
\newblock \bibinfo{title}{Neocortical network activity in vivo is generated
  through a dynamic balance of excitation and inhibition},
\newblock \bibinfo{journal}{Journal of Neuroscience} \bibinfo{volume}{26}
  (\bibinfo{year}{2006}) \bibinfo{pages}{4535--4545}.
\bibitem[{Rubin et~al.(2017)Rubin, Abbott, and Sompolinsky}]{rubin2017balanced}
\bibinfo{author}{R.~Rubin}, \bibinfo{author}{L.~Abbott},
  \bibinfo{author}{H.~Sompolinsky},
\newblock \bibinfo{title}{Balanced excitation and inhibition are required for
  high-capacity, noise-robust neuronal selectivity},
\newblock \bibinfo{journal}{Proceedings of the National Academy of Sciences}
  \bibinfo{volume}{114} (\bibinfo{year}{2017}) \bibinfo{pages}{E9366--E9375}.
\bibitem[{Sporns(2012)}]{sporns2012discovering}
\bibinfo{author}{O.~Sporns}, \bibinfo{title}{Discovering the human connectome},
  \bibinfo{publisher}{MIT press}, \bibinfo{year}{2012}.
\bibitem[{Cook et~al.(2019)Cook, Jarrell, Brittin, Wang, Bloniarz, Yakovlev,
  Nguyen, Tang, Bayer, Duerr et~al.}]{cook2019whole}
\bibinfo{author}{S.~J. Cook}, \bibinfo{author}{T.~A. Jarrell},
  \bibinfo{author}{C.~A. Brittin}, \bibinfo{author}{Y.~Wang},
  \bibinfo{author}{A.~E. Bloniarz}, \bibinfo{author}{M.~A. Yakovlev},
  \bibinfo{author}{K.~C. Nguyen}, \bibinfo{author}{L.~T.-H. Tang},
  \bibinfo{author}{E.~A. Bayer}, \bibinfo{author}{J.~S. Duerr}, et~al.,
\newblock \bibinfo{title}{Whole-animal connectomes of both caenorhabditis
  elegans sexes},
\newblock \bibinfo{journal}{Nature} \bibinfo{volume}{571}
  (\bibinfo{year}{2019}) \bibinfo{pages}{63--71}.
\bibitem[{Bates et~al.(2020)Bates, Schlegel, Roberts, Drummond, Tamimi,
  Turnbull, Zhao, Marin, Popovici, Dhawan et~al.}]{bates2020complete}
\bibinfo{author}{A.~S. Bates}, \bibinfo{author}{P.~Schlegel},
  \bibinfo{author}{R.~J. Roberts}, \bibinfo{author}{N.~Drummond},
  \bibinfo{author}{I.~F. Tamimi}, \bibinfo{author}{R.~G. Turnbull},
  \bibinfo{author}{X.~Zhao}, \bibinfo{author}{E.~C. Marin},
  \bibinfo{author}{P.~D. Popovici}, \bibinfo{author}{S.~Dhawan}, et~al.,
\newblock \bibinfo{title}{Complete connectomic reconstruction of olfactory
  projection neurons in the fly brain},
\newblock \bibinfo{journal}{BioRxiv}  (\bibinfo{year}{2020}).

\end{thebibliography}

\end{document}